\documentclass{article}

\usepackage{amsmath}
\usepackage{amssymb}
\usepackage{mathrsfs}
\usepackage{mathtools}
\usepackage{graphicx}
\usepackage{booktabs}
\usepackage[table]{xcolor}
\usepackage{wrapfig}
\usepackage[font=small]{caption}
\usepackage{xspace}
\usepackage{enumitem}
\usepackage{cleveref}
\usepackage{authblk}
\usepackage[margin=1in]{geometry}
\crefname{equation}{eq.}{eqs.}
\Crefname{equation}{Eq.}{Eqs.}
\newcommand{\myParagraph}[1]{%
  \textbf{#1.}\xspace%
}

\usepackage[dvipsnames]{xcolor}

\usepackage{multirow}

\usepackage[numbers,sort&compress]{natbib}
\usepackage{multibib}
\newcites{app}{References}

\providecommand{\bestcell}[1]{\cellcolor{green!20}#1}
\providecommand{\secondcell}[1]{\cellcolor{yellow!25}#1}

\title{Learning Adaptive Solvers for Distributed Factor Graph Optimization on Matrix Lie Groups}
\author[1]{Jaeho Shin}
\author[1]{Maani Ghaffari}
\author[1]{Yulun Tian}

\affil[1]{University of Michigan}
\date{}
\begin{document}
\maketitle
\vspace{-0.3in} % Tighten the title-to-teaser spacing.

%===============================================================================

\begin{abstract}
Modern robotic perception increasingly involves large-scale geometric optimization problems distributed across multiple robots or sessions. However, existing distributed solvers often depend on brittle hand tuning and primarily target rigid body pose graphs. To address this, we present DeepCORD, a learning-augmented framework for distributed factor graph optimization on general matrix Lie groups. By unfolding a parallel and accelerated Riemannian optimizer into differentiable iterations, DeepCORD learns a self-supervised feedback policy that dynamically adapts solver parameters according to the optimization phase and communication status. The resulting method enables adaptive distributed optimization over matrix Lie groups under both synchronous and asynchronous communication regimes. Extensive experiments on real-world $\mathrm{SE}$(3) pose graph optimization and $\mathrm{SL}$(4) projective submap alignment show that our method achieves lower objective values than existing distributed baselines on most benchmarks across realistic operating scenarios.
\end{abstract}

% \keywords{Learning to optimize, matrix Lie groups, multi-robot systems} 

%===============================================================================

% \YT{I thought more about this. ``Synchronization on Matrix Lie Groups'' is mathematically precise, but might be too optimization-ish/theoretical for the CoRL audience. I suggest we start from factor graph optimization over matrix Lie groups and use this for the majority of the introduction. Then we can specialize to synchronization problems when describing our contributions.}

% \begin{figure}[h]
%     \centering
%     \includegraphics[width=1\columnwidth]{fig1_sl4_alignment.pdf}
%     \caption{}
%     \label{fig:sl4}
% \end{figure}

\begin{figure}[h]
    \centering
\includegraphics[width=0.8\columnwidth]{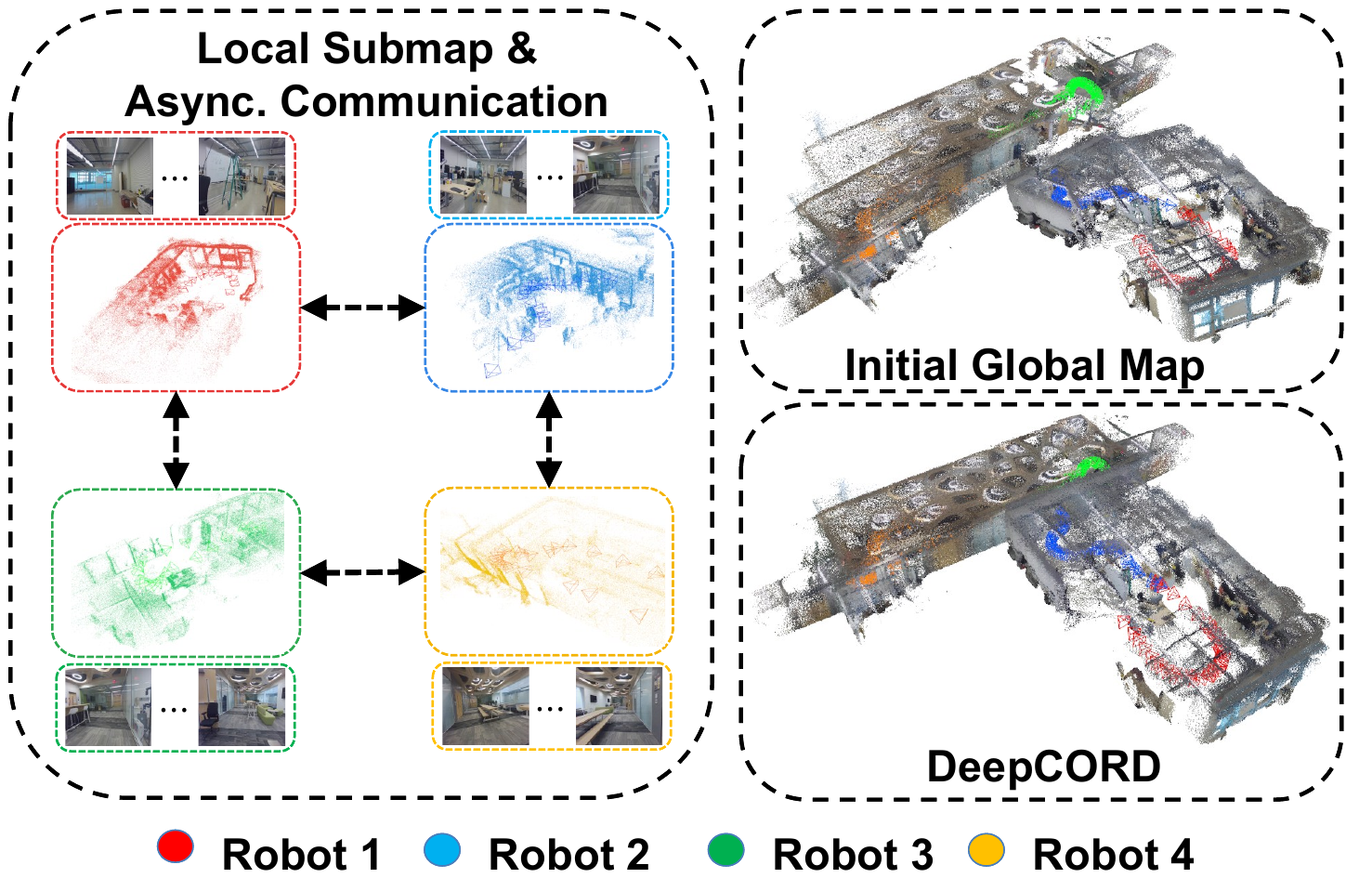}
    \caption{DeepCORD aligns feed-forward SLAM submaps from four robots through distributed $\mathrm{SL}(4)$ optimization, recovering a globally consistent map over a custom dataset spanning approximately 100 m.}
    \label{fig:SL4_result}
\end{figure}

\section{Introduction}

Modern robotic perception backends increasingly require solving large-scale geometric optimization problems, where many noisy measurements must be fused into a globally consistent spatial estimate. 
Factor graphs~\cite{dellaert2017factor,slam-handbook} provide a common abstraction for these problems and support optimization over different matrix Lie groups, including rigid motions for pose graph SLAM~\cite{rosen_se_sync_2019}, similarity transforms for monocular reconstruction~\cite{strasdat2010scale}, and 3D homographies for recent systems based on geometric foundation models \cite{maggio2026vggt,maggio2026vggt2}.
While centralized nonlinear solvers remain powerful when full graph access is available, many multi-robot, multi-device, and multi-session systems naturally produce factor graphs that are partitioned across robots or devices \cite{lajoie_door-slam_2020,tian2022kimera,lajoie_swarm-slam_2024}. In such settings, optimization must respect data locality, limited bandwidth, and potentially asynchronous communication.

Recent distributed solvers address this deployment setting by decomposing the backend across robots or graph partitions. 
Methods based on distributed optimization~\cite{tian_asynchronous_2020,fan_majorization_2024,mcgann_asynchronous_2024,shin2026distributed} and probabilistic message passing~\cite{murai_robot_2024,murai2024distributed,liu2025distributed} demonstrate that optimization can be achieved via local computation with periodic or asynchronous communication.
Nevertheless, the practical performance of these solvers is currently highly sensitive to manually selected solver parameters, such as step sizes, penalty parameters, damping coefficients, and restart rules.
Such sensitivity makes deployment brittle as tuned parameter settings fail to transfer across graph topology, noise level, or communication regime.
Further, existing distributed solvers  predominantly focus on rigid body SE(3) pose graphs, while recent perception systems increasingly involve optimization over broader matrix Lie groups.

\myParagraph{Contributions}
Inspired by learning-to-optimize (L2O) frameworks~\cite{chen2022learning}, we propose DeepCORD, a learning-augmented framework for distributed factor graph optimization on matrix Lie groups. 
DeepCORD builds on CORD~\cite{shin2026distributed}, a state-of-the-art distributed solver based on second-order Riemannian dynamics, and unfolds its iterations into a differentiable computation graph. Rather than learning black-box state updates, DeepCORD learns a local feedback policy that predicts adaptive solver parameters from local optimization context. 
The resulting learned update preserves the distributed structure while adapting its dynamics across problem instances, optimization stages, and communication regimes. 
% To the best of our knowledge, this is the first work that extends L2O to distributed factor graph optimization. 
In summary, 
\begin{itemize}[ leftmargin=*]
    \item We present \textbf{DeepCORD}, a learning-augmented distributed optimizer on matrix Lie groups.

    \item We design a self-supervised approach to train DeepCORD via deep unfolding, by minimizing unrolled objectives without relying on supervision from optimal solutions.

    \item We evaluate DeepCORD on an extensive collection of real-world pose graph optimization (PGO) and projective submap alignment problems.
    DeepCORD outperforms hand-tuned baselines across communication regimes, achieving the best results on 21 out of 26 PGO benchmarks and all three projective alignment datasets.
    
    % \item We present \textbf{DeepCORD}, a learning-augmented framework for
    % distributed optimization over matrix Lie groups.
    % Via self-supervised training, {DeepCORD} learns local policies to adapt optimization dynamics across problem instances,
    % optimization stages, and communication regimes;

    % \item We validate {DeepCORD} on two representative matrix Lie group
    % synchronization problems: (i) distributed pose graph
    % optimization over $\mathrm{SE}(3)$, and (ii) projective submap alignment over $\mathrm{SL}(4)$ manifolds integrated with recent geometric foundation model front-end. 
    % Extensive experiments verify improved objective reduction over distributed solvers across benchmark PGO and multi-session reconstruction scenarios, achieving best performance on 13 out of 14 synchronous SE(3) PGO benchmarks.%\YT{Inclide one or two quantitative numbers as highlight.}
\end{itemize}

\section{Related Work}\label{sec:RelatedWork}

\myParagraph{Factor Graph Optimization} 
Factor graphs support optimization over different matrix Lie groups, including rotation averaging \cite{singer2011angular, dellaert2020shonan}, pose graph optimization (PGO) \cite{rosen_se_sync_2019}, and scale-drift correction with similarity transforms \cite{strasdat2010scale}. 
Recently, VGGT-SLAM~\citep{maggio2026vggt,maggio2026vggt2} uses the special linear group $\mathrm{SL}(4)$ to represent projective transformations in homogeneous coordinates for aligning feed-forward SLAM submaps under projective ambiguity.
The standard approach to factor graph optimization uses centralized nonlinear least-squares solvers~\cite{gtsam, ceres, kummerle2011g}.
Recently, \textbf{distributed solvers} have gained increasing attention.
% with distributed PGO receiving particular attention due to its application in multi-robot SLAM \cite{lajoie_door-slam_2020,tian2022kimera,lajoie_swarm-slam_2024}.
DC2-PGO~\citep{tian_distributed_2021} solves distributed PGO via a sparse semidefinite relaxation.
Other methods improve scalability or robustness through accelerated
majorization minimization~\citep{fan_majorization_2024,fan2025daba}, asynchronous parallel
gradient descent~\citep{tian_asynchronous_2020}, 
on-manifold consensus ADMM~\citep{mcgann_asynchronous_2024,mcgann_imesa_2024},
and overlapping domain decomposition~\citep{sonawalla2026overlapping}. A complementary class of solvers is based on Gaussian belief propagation (GBP), with applications to distributed localization and sensor calibration~\citep{murai_robot_2024,murai2024distributed,liu2025distributed}. %\YT{Rewrite this sentence. Many-device localization and distributed PGO is the same thing in these papers.}
Despite the progress, the practical performance of existing distributed solvers depends heavily on parameter tuning. DeepCORD addresses this challenge through a learned policy, enhancing empirical robustness while preserving the principled geometric updates from the underlying solver.

\textbf{Learning to optimize (L2O)} is an optimization framework leveraging neural networks to accelerate iterative solvers through learned updates, initializations, or algorithmic parameters~\citep{amos2023tutorial}. One line of work trains neural networks to directly regress the optimization update at each iteration~\citep{andrychowicz2016learning,li2016learning}. A complementary approach learns to predict a warm start, which is then refined by a model-based optimizer~\citep{sambharya2023end,sambharya2024learning}. Model-based L2O methods preserve the update rules of the underlying solvers and learn iteration-dependent parameters often via deep unfolding~\citep{gregor2010learning}. Deep unfolding has been applied to tune first-order and consensus-based quadratic programming (QP) solvers~\citep{ichnowski2021accelerating,saravanos2025deep}, nonlinear programming solvers~\citep{oshin2026deep}, and distributed ADMM~\citep{noah2024distributed,doerks2025learning}.  %\YT{Rewrite this sentence: first-order QP and consensus QP have strong overlap and are better merged}. 
In the context of geometric estimation, L2O has also been applied to two-view reconstruction~\citep{clark2018learning}, rotation averaging~\citep{thorpe2020rotation,purkait2020neurora} and PGO~\citep{li2021pogo,kourtzanidis2023rl,ghanta2025policies}. However, none of these works has addressed the challenging setting of distributed factor graph optimization on general matrix Lie groups under asynchronous communication. 
DeepCORD bridges this gap by learning an optimization solver that dynamically adapts parameters while supporting broader groups beyond rotations and poses.
\section{Problem Formulation}

In this work, we consider distributed factor graph optimization problems with pairwise relative measurements between group-valued variables. 
This class of factor graphs forms the backbone for modern SLAM and 3D reconstruction systems \citep{dellaert2017factor,slam-handbook}.
Formally, let $G \subseteq \mathrm{GL}(n)$ be a matrix Lie group with Lie algebra $\mathfrak{g}$.
Let $\mathcal{G} = (\mathcal{V}, \mathcal{E})$ represent a connected factor graph, where each node $u \in \mathcal{V}$ represents an optimization variable $X_u \in G$.
Each edge $(u,v) \in \mathcal{E}$ represents a noisy pairwise measurement $Z_{uv} \approx X_u^{-1} X_v$.
Given a cost term corresponding to each edge $c_{uv}: G \times G \to \mathbb{R}_{\geq 0}$, 
our goal is to solve the following optimization problem over all nodes $X \coloneq \{X_u\}_{u \in \mathcal{V}}$,
\begin{equation}
    \min_{X \in G^{|\mathcal{V}|}} \sum_{(u,v) \in \mathcal{E}} c_{uv}(X_u Z_{uv}, X_v).
    \label{eq:problem}
\end{equation}

\begin{wrapfigure}{r}{0.3\textwidth}
    \vspace{-1.0em}
    \centering
    \includegraphics[width=\linewidth]{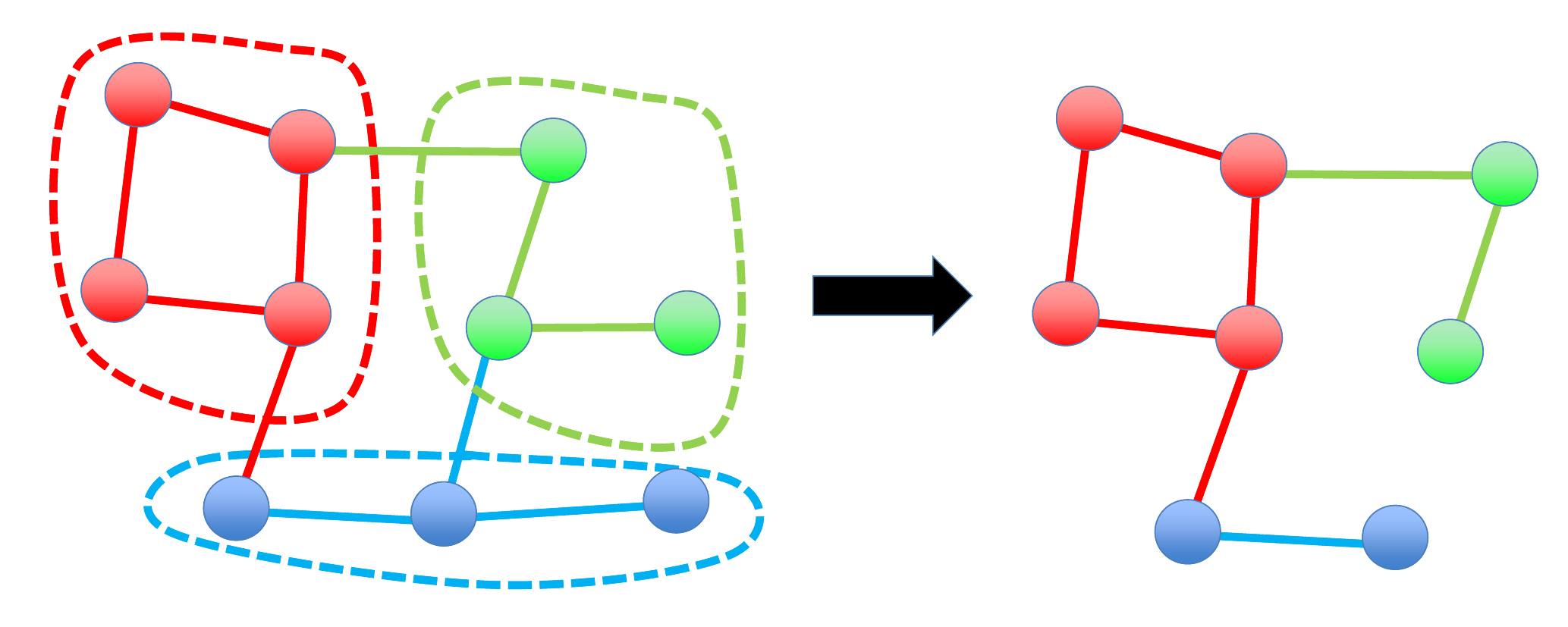}
    \caption{(Left) Node and edge partitioning in distributed optimization. Colors indicate different robots. (Right) The augmented graph for the red robot. }
    \label{fig:graph}
    \vspace{-2.0em}
\end{wrapfigure} In the distributed setting, the nodes are partitioned across a set of robots $\mathcal A=\{1,\ldots,N\}$. Robot $i\in\mathcal A$ owns a subset of variables $\mathcal V_i\subseteq\mathcal V$ and maintains the corresponding local states $X^i \coloneq \{X_\ell\}_{\ell\in\mathcal V_i}$. Each robot only optimizes the variables in its own subgraph.
In addition, we create a disjoint partitioning of the edge (measurement) set $\mathcal{E} = \mathcal{E}_1 \uplus \hdots \uplus \mathcal{E}_N$, by arbitrarily assigning each inter-robot edge to one of the incident robots; see Fig.~\ref{fig:graph}.
Given the edge partitioning, each robot has a corresponding local cost defined by summing all edge costs it owns,
\begin{equation}
    \mathcal{C}^i(X) \coloneq \sum_{(u,v) \in \mathcal{E}_i} c_{uv}(X_u Z_{uv}, X_v).
\end{equation}
\begin{wrapfigure}{r}{0.50\textwidth}
    \centering
    \includegraphics[width=\linewidth]{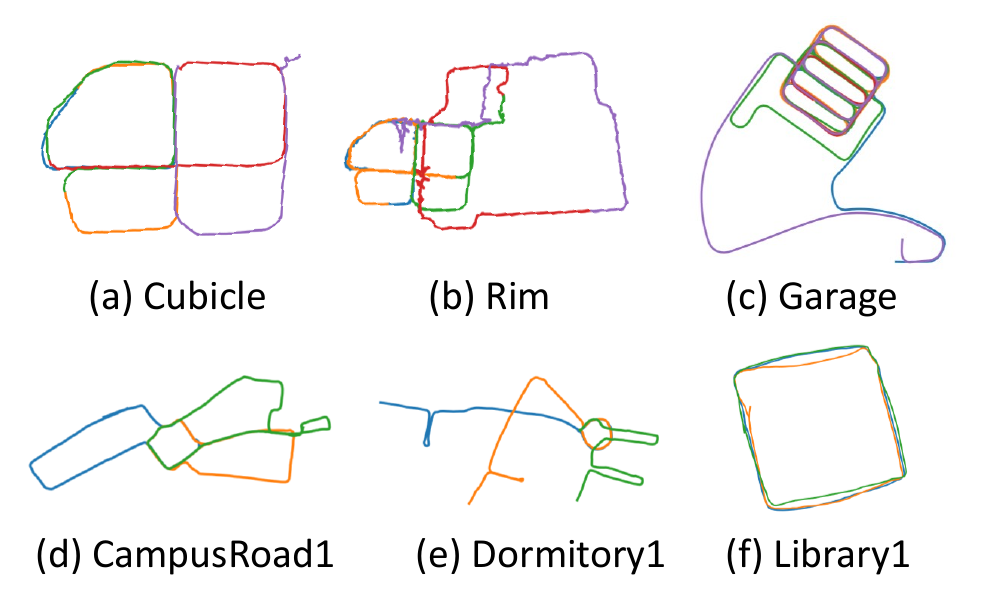}
    \caption{Distributed PGO solved by the proposed method. Each color denotes a different robot.}
    \label{fig:Trajectory}
\end{wrapfigure}
The overall cost in \eqref{eq:problem} is the sum of all local costs. Importantly, for each robot $i$ to evaluate $\mathcal{C}^i$, it only needs access to its own nodes $X^i$ and a subset of boundary nodes from neighboring robots.
In this work, we study the following two important instantiations of the optimization problem in \eqref{eq:problem}.

\myParagraph{Pose Graph Optimization (PGO) over $\mathrm{SE}(3)$} 
State-of-the-art multi-robot SLAM systems use PGO to correct
accumulated odometry drift and improve global consistency of trajectory estimation \cite{lajoie_door-slam_2020,tian2022kimera,lajoie_swarm-slam_2024}; see Fig.~\ref{fig:Trajectory}. In PGO, each node variable  \(X_u\) models a robot pose in the global frame as an element of the rigid body motion group $\mathrm{SE}(3) = \{ [R, t; \mathbf{0}^\top, 1] \in \mathbb{R}^{4 \times 4} \mid R\in\mathrm{SO}(3),\ t\in\mathbb R^3 \}$.
The associated Lie algebra $\mathfrak{se}(3) = \{ [\Omega, \rho; \mathbf{0}^\top, 0] \in \mathbb{R}^{4 \times 4} \mid \Omega \in \mathfrak{so}(3), \rho \in \mathbb{R}^3 \}$ is 6-dimensional.
Following \cite{rosen_se_sync_2019, fan_majorization_2024, tian_distributed_2021, sonawalla2026overlapping,shin2026distributed}, we define the edge cost using the squared chordal distance $c_{uv} = \left\|X_u Z_{uv} - X_v\right\|^2_{\Omega_{uv}}$ where the weighted norm is defined as $\| A \|_{\Omega}^2 \coloneq \mathrm{tr}\left( A \Omega A^\top \right)$. $\Omega = \mathrm{blkdiag}(\omega_{R} I_3, \omega_{t})$ is the precision matrix with scalar weights $\omega_R, \omega_t > 0$ for rotation and translation, respectively.

\myParagraph{Projective Submap Alignment over $\mathrm{SL}(4)$} 
Recent geometric foundation models such as \citep{wang2025vggt} directly output 3D reconstruction from uncalibrated images.
To improve global reconstruction, state-of-the-art SLAM systems \citep{maggio2026vggt,maggio2026vggt2} perform submap-level projective alignment.
% Projective submap alignment arises in recent feed-forward SLAM systems \citep{maggio2026vggt,maggio2026vggt2} based on geometric foundation model~\citep{wang2025vggt}, where local reconstructions are subject to projective ambiguities.
In this problem, each node variable \(X_u\) represents a 3D homography that maps local poses and points in a submap to a global reference frame, and is modeled as an element of the special linear group $\mathrm{SL}(4)=\{X\in\mathbb R^{4\times4}\mid \det(X)=1\}$ to account for projective ambiguity. The associated Lie algebra $\mathfrak{sl}(4)=\{A\in\mathbb R^{4\times4}\mid \mathrm{tr}(A)=0\}$ is 15-dimensional. 
Following VGGT-SLAM \citep{maggio2026vggt,maggio2026vggt2}, 
we define the alignment error using the geodesic residual $c_{uv} =\left\|\log\!\left(Z_{uv}^{-1}X_u^{-1}X_v\right) \right\|^2_{\Omega_{uv}},$ where \(\log:\mathrm{SL}(4)\rightarrow\mathfrak{sl}(4)\) denotes the logarithm mapping.
The weighted norm is defined as
    $\|A\|_\Omega^2
    \coloneq
    \psi(A)^\top \Omega \psi(A)
$ where $\psi(A)\in\mathbb R^{15}$ 
denotes the coordinate representation of $A$ in a fixed basis of \(\mathfrak{sl}(4)\) and $\Omega = \omega I_{15}$ is the precision matrix with positive scalar weight $\omega$.

\section{Proposed Method}
DeepCORD arises from unfolding the iterations of CORD~\cite{shin2026distributed}, a state-of-the-art distributed solver, to a learning-augmented framework.
In the following, we will review necessary background, discuss the neural network architecture, 
and then present the self-supervised training approach.
Details about the network architecture, feature definitions, and training setup are deferred to the Appendix.

\myParagraph{Background: CORD \cite{shin2026distributed}}
CORD applies the Euler--Poincar\'{e} equation \cite{bloch1996euler} to solve distributed optimization over general matrix Lie groups $G$.
For each node in the factor graph $X_u \in G$, let $\xi_u \coloneqq (X_u^{-1}\dot X_u) \in \mathfrak{g}$ denote its body velocity. 
% For robot $i$, let $\xi^i \in \mathfrak{g}^{|\mathcal{V}_i|}$ be the stacked body velocity of its local variables.
In addition, let $\xi \in \mathfrak{g}^{|\mathcal{V}|}$ denote the stacked body velocity over all variables in the factor graph.
CORD formulates a continuous time system that governs the evolution of the optimization variables,
\begin{equation}
     M\dot{\xi} =  - \operatorname{grad}_{X} \mathcal{C}(X) 
    -D\xi
    +\text{ad}^*_{\xi}(M\xi) - \dot{M}{\xi}.
    \label{eq:cord_continuous}
\end{equation}
Above, $M\xi$ is a momentum term, $-\operatorname{grad}_{X} \mathcal{C}(X)$ is the body-trivialized negative gradient of the cost function that serves as the potential energy, $-D\xi$ is a damping term, and $\operatorname{ad}^{*}_{\xi}(M\xi)$ is the co-adjoint term that accounts for Lie-group geometry. The $\dot{M}\xi$ term accounts for possibly time-varying mass matrix $M$.
% the dot operator denotes the derivative with respect to the algorithmic time. 
In \cite{shin2026distributed}, it is shown that with suitable choices of algorithm parameters, solving \eqref{eq:cord_continuous} recovers a local minimizer to the optimization problem.
In practice, CORD solves \eqref{eq:cord_continuous} via a distributed integration scheme in which $M=\operatorname{blkdiag}(M^1,\ldots,M^N)$ and
$D=\operatorname{blkdiag}(D^1,\ldots,D^N)$ are block-diagonal matrices. Specifically, CORD chooses $M^i = m H^i$ and $D^i = d H^i$, where $m, d > 0$ are scalar parameters constant across all robots and iterations, and $H^i$ is the diagonal block of the Gauss-Newton Hessian approximation corresponding to robot $i$.
This design natually decomposes the integration across robots:
at every discretized iteration $k$, each robot $i$ computes the following update in parallel,
\begin{align}
    &\dot{\xi}^{i}_{k}
    =
    (mH^{i}_{k})^{-1}
    \big(
    -\operatorname{grad}_{X_k^{i}} \mathcal{C}(X_{k})
    - dH^{i}_{k}\xi^{i}_{k}
    + \operatorname{ad}^{*}_{\xi^{i}_{k}}(mH^{i}_{k}\xi^{i}_{k})
    - m\dot{H}^{i}_{k}\xi^{i}_{k}
    \big), \label{eq:cord_discrete_acc}\\
    &\xi^{i}_{k+1}
    =
    \xi^{i}_{k}
    +
    \dot{\xi}^{i}_{k}\Delta t, 
    \quad
    X^{i}_{k+1}
    =
    X^{i}_{k}
    \exp\!\left(\xi^{i}_{k+1}\Delta t\right). \label{eq:cord_discrete_exp}
\end{align}
\Cref{eq:cord_discrete_acc} solves the local acceleration term $\dot{\xi}^{i}_{k}$ according to \eqref{eq:cord_continuous}, and \eqref{eq:cord_discrete_exp} applies a semi-implicit Euler step to update the body velocity and the variable using a step size $\Delta t > 0$.
% Because $\operatorname{grad}_{X_k^i} \mathcal{C}(X_k)$ depends only on robot $i$'s intra-robot factors and inter-robot factors incident to its neighbors, each iteration 
% \eqref{eq:cord_discrete_acc}-\eqref{eq:cord_discrete_exp} requires only one round of communication to exchange the neighboring states needed to evaluate the inter-robot residuals.

\begin{figure}[t]
    \centering
    \includegraphics[width=1\columnwidth]{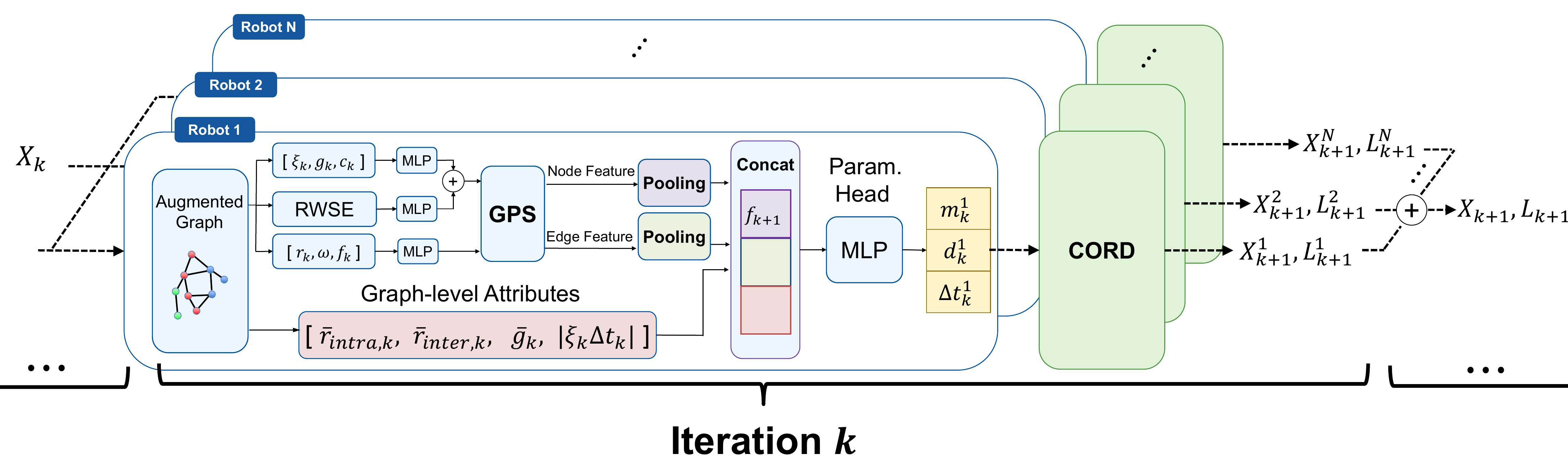}
    \caption{Proposed architecture of the adaptive local feedback policy $\pi_\theta$ in DeepCORD.}
    \label{fig:architecture}
\end{figure}

\myParagraph{DeepCORD: Learning Adaptive Parameters via Local Feedback Policy} 
In practice, the performance of CORD depends strongly on the manually tuned mass $m$, damping $d$, and step size $\Delta t$ parameters.
% Despite the improvements demonstrated in \cite{shin2026distributed}, CORD relies on manually tuned, constant parameter settings for the momentum $m$, damping $d$, and step size $\Delta t$.
% Further, empirical performance of the solver is highly sensitive to the values of these parameters.
Our central insight is to replace manual tuning with an adaptive local feedback policy $\pi_\theta$ as shown in Fig.~\ref{fig:architecture}. The weights $\theta$ are shared across all robots and problem instances, enabling generalization across graph sizes, topologies, and communication regimes.
At every iteration $k$, each robot $i$ applies this shared policy to local information to adjust its own parameters,
\begin{equation}
    m^i_k,\ d^i_k,\ \Delta t^i_k
    =
    \pi_\theta( \mathcal{G}^i_k), 
    \label{eq:deepcord_policy}
\end{equation}
and uses these updated parameters in the next iteration according to \eqref{eq:cord_discrete_acc}-\eqref{eq:cord_discrete_exp}.
The input to the policy is a local \emph{augmented graph} $\mathcal{G}^i_k$ that encodes optimization context.
Specifically, $\mathcal{G}^i_k$ extends the robot's own subgraph by two hops to include the most recently received boundary nodes from neighboring robots; see Fig.~\ref{fig:graph}.
% Derived from the current state estimate $X^k$, the attributes of $\mathcal{G}^i_k$ capture the ongoing optimization status, such as local residuals and gradients. We adopt this two-hop formulation for our main evaluations, with a corresponding ablation study on neighborhood size provided in \blue{Appendix~XXX}. See Fig.~\ref{fig:graph} for visualization of augmented graph.
We attach compact node-, edge-, and graph-level attributes to $\mathcal{G}^i_k$.
Specifically, node attributes contain body velocity $\xi_k$, local gradient $g_k$, random-walk structural encoding (RWSE) \cite{dwivedi2021graph}, and a communication feature $c_k$ capturing delay statistics.
Edge attributes contain edge residual $r_k$, precision vector $\omega$, and a robot-level global context vector $f_k$ from previous iteration.
Graph-level attributes contain intra-robot residual norm \(\bar{r}_{\text{intra},k}\), the inter-robot residual norm \(\bar{r}_{\text{inter},k}\), the robot gradient norm \(\bar{g}_k\), and the previous update norm $|\xi_k \Delta t_k|$.
The exact attribute definitions are provided in the Appendix.

At every iteration, the policy $\pi_\theta$ processes the graph attributes using a graph encoder followed by a MLP parameter head.
Node and edge attributes are first embedded by small MLPs and then processed by GPS layers \citep{rampavsek2022recipe}.
The results are pooled and concatenated with graph-level attributes before being passed to the parameter head to predict $m^i_k$, $d^i_k$, and $\Delta t^i_k$.
Notably, DeepCORD retains the communication efficiency of CORD and each iteration uses a single round of communication.

\myParagraph{Self-Supervised Learning via Deep Unfolding}
We train the policy $\pi_\theta$ in a self-supervised manner by
minimizing the cost function evaluated along the unrolled iterates.
Let $\mathcal{S} = \{\zeta_j\}_{j=1}^{H}$ denote a set of
$H$ training problem instances.
For each problem $\zeta_j$, let $X_k \equiv X_k(\zeta_j;\theta)$
denote the unrolled solution after $k$ DeepCORD iterations where we drop the problem index $j$ to simplify notation.
For each robot $i$, we normalize its local cost by intra-robot edge count,
$\bar{\mathcal{C}}^i(X_k) = \mathcal{C}^i(X_k)/(|\mathcal{E}_{\mathrm{intra}}^i| + \epsilon)$
which we find to improve training performance.
In addition, we introduce two regularization terms.
First, the \emph{monotonicity} term
\(B^i_k \coloneq \max(0,\bar {\mathcal{C}}^i(X_k)-\bar{\mathcal{C}}^i(X_{k-1}))\) penalizes cost increase for consecutive iterations.
Second, motivated by the observation in \cite{shin2026distributed}, the \emph{damping} term \(D^i_k \coloneq \max\{0,\, d^i_k-d_{k-1}^{i}\exp[-\beta/(t^i_{k-1}+\tau)]\}\) encourages decaying damping coefficient $d$ where $\beta,\tau$ are constants and $t^i_{k-1}$ is the cumulative integration time.
The loss for each problem $\zeta_j$ averages the normalized cost over all robots along $K=50$ unrolled iterations together with the regularization terms, 
\begin{equation}
  \mathcal{L}^{(j)}
  \;\coloneq\;
  \frac{1}{N_j K}
  \sum_{k=1}^{K}\sum_{i=1}^{N_j}
  \Big[\,
      \bar{\mathcal{C}}^{\,i}(X_k)
      \;+\; \lambda_b\, B^{\,i}_{k}
      \;+\; \lambda_d\, D^{\,i}_{k}
  \,\Big],
  \label{eq:deepcord-loss}
\end{equation}
where $N_j$ is the number of robots in problem $j$.
The overall training loss is the average over all training problems 
$\mathcal{L} \coloneq \sum_{j=1}^H \mathcal{L}^{(j)}/H$. 
During training, recall from \eqref{eq:cord_discrete_acc} that each DeepCORD iteration requires inverting the local Hessian $H^i_k$.
For numerical efficiency, we implement sparse preconditioned conjugate gradient and backpropagate through the linear solve via implicit differentiation as in ~\cite{saravanos2025deep}.
For training stability, we additionally unroll a simplified ODE that
omits the co-adjoint and time-dependent mass terms from Eq.~\eqref{eq:cord_discrete_acc}. We provide an additional discussion of the simplified dynamics in the {Appendx}.

\myParagraph{Training Setup}
\label{subsec:training_data}
We train a separate DeepCORD model for $\mathrm{SE}(3)$ and $\mathrm{SL}(4)$ optimization.
To enable generalization, we utilize a mixture of 1,000 synthetic multi-robot pose graphs spanning five topologies (\texttt{grid}, \texttt{sphere}, \texttt{torus}, \texttt{helix}, \texttt{random walk}) and 1,440 real-trajectory sequences from public datasets~\cite{mcgann2025cosmo,albin2025cu,tian2022kimera,carlevaris2016university,reinke2022locus}. Across the training instances, the number of robots ranges from 2 to 8, with each graph comprising approximately 350–500 poses. For projective submap alignment over $\mathrm{SL}(4)$, we utilize the same graph setup as in PGO but generate relative measurements following the VGGT-SLAM 2.0~\cite{maggio2026vggt2} measurement model.
Details about data curation and training configuration are discussed in the Appendix.

\section{Experiments}
\label{sec:experiments}

In this section, we validate DeepCORD on standard PGO over $\mathrm{SE}(3)$ and projective submap alignment over $\mathrm{SL}(4)$. %\YT{In problem formulation, we defined PGO and project submap alignment. It's good practice to stick to these terminologies consistently throughout the paper.}
Our experiments demonstrate that DeepCORD achieves the best cost on 11 out of 13 synchronous instances and 10 out of 13 asynchronous instances, and predicts adaptive parameters that generalize to larger datasets and longer optimization iterations. 

\myParagraph{Experiment Setup}
\label{subsec:experiment_setup}
To test generalization, all problems in the experiments are not seen during training.
% \YT{Emphasize that these datasets are not seen during training.}
For PGO, we evaluate on standard benchmarks \cite{fan_majorization_2024, shin2026distributed} and the real-world multi-robot S3E dataset~\cite{feng2024s3e}; see Fig.~\ref{fig:Trajectory}.
% \blue{For S3E, we generate pose graphs using the LiDAR module of Swarm-SLAM \YT{Cite} and filter outlier edges using ground-truth information.}
For projective submap alignment, we use three TUM RGB-D sequences~\cite{sturm2012benchmark}: \texttt{fr1/room}, \texttt{fr2/desk}, and \texttt{fr3/household}, denoted as \texttt{TUM1}, \texttt{TUM2}, and \texttt{TUM3}. 
In addition, we collect a larger custom real-world dataset shown in Fig.~\ref{fig:SL4_result}.
Each sequence is divided into four subsequences with overlapping fields of view. 
We first run VGGT-SLAM \cite{maggio2026vggt} on each subsequence to obtain local subgraphs, and then construct inter-session loop closures using the loop closure module of \cite{maggio2026vggt}. 
% To further validate the practical capability, we also collect a real-world, floor-scale campus-building dataset and evaluate our method on the resulting multi-session projective submap alignment problem; see Fig.~\ref{fig:SL4_result}.
% \YT{Discuss the custom dataset.}
Experiments are conducted under both synchronous and asynchronous communication. 
For asynchronous communication, 
we follow \cite{shin2026distributed} and simulate random inter-robot packet delay between 0 and 10 steps and random packet drop rate of $10\%$. 
For the asynchronous setting, we perform 10 Monte Carlo runs for each method and report the mean value.

\myParagraph{Metrics and Baselines} \label{subsec:metrics_and_baselines} We report the cost values achieved by all methods as the main performance metric. Additional evaluation metrics regarding accuracy and efficiency are reported in the Appendix. Reference costs are obtained by the global solver SE-Sync \cite{rosen_se_sync_2019} for PGO and GTSAM \cite{gtsam} for projective submap alignment.
%\YT{Do we measure ATE and RMSE? Also, discuss the metrics for SL(4).}
For synchronous PGO, we compare against AMM-PGO \cite{fan_majorization_2024}, CORD~\cite{shin2026distributed}, and ROBO~\cite{sonawalla2026overlapping}. ROBO uses an overlap parameter of one, which means the method has access to 2-hop neighborhood information as in the proposed approach. Unless mentioned otherwise, all hyperparameters for the baselines are set to default values. For asynchronous PGO, we compare against Distributed Jacobi (DJ) and CORD, where DJ employs a preconditioned gradient descent update rule with a robot-level block-diagonal Hessian as in \cite{tian_asynchronous_2020}. 
For projective submap alignment over $\mathrm{SL}(4)$, we focus on the same asynchronous settings with DJ and CORD as baselines.

% We compare against AMM-PGO \cite{fan_majorization_2024}, RBCD \cite{tian_distributed_2021}, and ROBO \cite{sonawalla2026overlapping} with an overlap setting of $\omega=1$ which leverages 2-hop neighborhood information similar to DeepCORD, and CORD utilizing constant parameters across all dataset with same communication and initialization scheme. 

\subsection{Pose Graph Optimization over $\mathrm{SE}(3)$}

% For PGO, we consider two strategies to initialize distributed optimization: chordal initialization \cite{fan_majorization_2024} and coarse alignment \cite{tian2022kimera}.
% In coarse alignment, we first optimize each robot's individual pose graph, and then obtain a multi-robot pose graph via a robot-level spanning tree.
% For our main result, we report the chordal initialization results and  coarse alignment results are provided at the appendix.

We initialize all methods using the chordal initialization scheme as in \cite{fan_majorization_2024,shin2026distributed}. Results using alternative initialization are presented in the Appendix.

\begin{table*}[t]
    \setlength{\tabcolsep}{0.24em}
    \centering
    \caption{Synchronous and asynchronous $\mathrm{SE}(3)$ results from chordal initialization after 100 distributed iterations. The best and second-best methods within each setting are highlighted in green and yellow, respectively.}
    \resizebox{\textwidth}{!}{%
    \renewcommand{\arraystretch}{1.5}
    \begin{tabular}{|c||c|c|c||c|c|c|c||c|c|c|}
        \hline
        \multirow{2}{*}{Dataset}
        & \multirow{2}{*}{\# Nodes}
        & \multirow{2}{*}{\# Edges}
        & \multirow{2}{*}{Ref.}
        & \multicolumn{4}{c||}{Synchronous}
        & \multicolumn{3}{c|}{Asynchronous} \\
        \cline{5-11}
        & & & 
        & ROBO & AMM & CORD & DeepCORD
        & CORD & DJ & DeepCORD \\
        \hline

        {\sf Sphere}
        & $2500$ & $4949$ & $1.687\mathrm{e}{3}$
        & \bestcell{$1.687\mathrm{e}{3}$}
        & \bestcell{$1.687\mathrm{e}{3}$}
        & \bestcell{$1.687\mathrm{e}{3}$}
        & \bestcell{$1.687\mathrm{e}{3}$}
        & \bestcell{$1.689\mathrm{e}{3}$}
        & $1.692\mathrm{e}{3}$
        & \bestcell{$1.689\mathrm{e}{3}$} \\
        \hline

        {\sf Torus}
        & $5000$ & $9048$ & $2.423\mathrm{e}{4}$
        & \bestcell{$2.423\mathrm{e}{4}$}
        & \bestcell{$2.423\mathrm{e}{4}$}
        & \bestcell{$2.423\mathrm{e}{4}$}
        & \bestcell{$2.423\mathrm{e}{4}$}
        & \bestcell{$2.423\mathrm{e}{4}$}
        & \secondcell{$2.424\mathrm{e}{4}$}
        & \bestcell{$2.423\mathrm{e}{4}$} \\
        \hline

        {\sf Grid}
        & $8000$ & $22236$ & $8.432\mathrm{e}{4}$
        & \bestcell{$8.432\mathrm{e}{4}$}
        & \bestcell{$8.432\mathrm{e}{4}$}
        & \bestcell{$8.432\mathrm{e}{4}$}
        & \bestcell{$8.432\mathrm{e}{4}$}
        & \secondcell{$8.433\mathrm{e}{4}$}
        & \secondcell{$8.433\mathrm{e}{4}$}
        & \bestcell{$8.432\mathrm{e}{4}$} \\
        \hline

        {\sf Cubicle}
        & $5750$ & $16869$ & $7.171\mathrm{e}{2}$
        & $7.194\mathrm{e}{2}$
        & $7.180\mathrm{e}{2}$
        & \secondcell{$7.176\mathrm{e}{2}$}
        & \bestcell{$7.173\mathrm{e}{2}$}
        & \bestcell{$7.235\mathrm{e}{2}$}
        & $7.251\mathrm{e}{2}$
        & \bestcell{$7.235\mathrm{e}{2}$} \\ 
        \hline

        {\sf Rim}
        & $10195$ & $29743$ & $5.461\mathrm{e}{3}$
        & $5.513\mathrm{e}{3}$
        & $5.482\mathrm{e}{3}$
        & \secondcell{$5.475\mathrm{e}{3}$}
        & \bestcell{$5.471\mathrm{e}{3}$}
        & \bestcell{$5.736\mathrm{e}{3}$}
        & $5.785\mathrm{e}{3}$
        & \secondcell{$5.751\mathrm{e}{3}$} \\
        \hline

        {\sf Garage}
        & $1661$ & $6275$ & $1.263\mathrm{e}{0}$
        & $1.267\mathrm{e}{0}$
        & $1.268\mathrm{e}{0}$
        & \bestcell{$1.265\mathrm{e}{0}$}
        & \secondcell{$1.266\mathrm{e}{0}$}
        & \bestcell{$1.280\mathrm{e}{0}$}
        & \secondcell{$1.283\mathrm{e}{0}$}
        & \bestcell{$1.280\mathrm{e}{0}$} \\
        \hline

        {\sf S3E CampusRoad1}
        & $2348$ & $2675$ & $4.365\mathrm{e}{4}$
        & $4.379\mathrm{e}{4}$
        & \secondcell{$4.374\mathrm{e}{4}$}
        & $4.376\mathrm{e}{4}$
        & \bestcell{$4.373\mathrm{e}{4}$}
        & \bestcell{$4.394\mathrm{e}{4}$}
        & \secondcell{$4.395\mathrm{e}{4}$}
        & $4.397\mathrm{e}{4}$ \\
        \hline

        {\sf S3E CampusRoad2}
        & $3336$ & $4212$ & $1.018\mathrm{e}{5}$
        & \bestcell{$1.019\mathrm{e}{5}$}
        & \bestcell{$1.019\mathrm{e}{5}$}
        & \bestcell{$1.019\mathrm{e}{5}$}
        & \bestcell{$1.019\mathrm{e}{5}$}
        & \bestcell{$1.022\mathrm{e}{5}$}
        & \bestcell{$1.022\mathrm{e}{5}$}
        & \bestcell{$1.022\mathrm{e}{5}$} \\
        \hline

        {\sf S3E CampusRoad3}
        & $2314$ & $3504$ & $1.530\mathrm{e}{5}$
        & $1.767\mathrm{e}{5}$
        & \bestcell{$1.675\mathrm{e}{5}$}
        & \secondcell{$1.693\mathrm{e}{5}$}
        & $1.696\mathrm{e}{5}$
        & \bestcell{$2.136\mathrm{e}{5}$}
        & $2.145\mathrm{e}{5}$
        & \secondcell{$2.143\mathrm{e}{5}$} \\
        \hline

        {\sf S3E Dormitory1}
        & $2194$ & $2304$ & $2.045\mathrm{e}{4}$
        & $2.072\mathrm{e}{4}$
        & \secondcell{$2.056\mathrm{e}{4}$}
        & \secondcell{$2.056\mathrm{e}{4}$}
        & \bestcell{$2.052\mathrm{e}{4}$}
        & \secondcell{$2.373\mathrm{e}{4}$}
        & $2.386\mathrm{e}{4}$
        & \bestcell{$2.350\mathrm{e}{4}$} \\
        \hline

        {\sf S3E Library1}
        & $1801$ & $2049$ & $2.979\mathrm{e}{4}$
        & $3.032\mathrm{e}{4}$
        & \secondcell{$3.006\mathrm{e}{4}$}
        & $3.012\mathrm{e}{4}$
        & \bestcell{$2.990\mathrm{e}{4}$}
        & \secondcell{$3.065\mathrm{e}{4}$}
        & $3.067\mathrm{e}{4}$
        & \bestcell{$3.062\mathrm{e}{4}$} \\
        \hline

        {\sf S3E Library2}
        & $519$ & $623$ & $2.297\mathrm{e}{4}$
        & \secondcell{$2.298\mathrm{e}{4}$}
        & \bestcell{$2.297\mathrm{e}{4}$}
        & \secondcell{$2.298\mathrm{e}{4}$}
        & \bestcell{$2.297\mathrm{e}{4}$}
        & \secondcell{$2.305\mathrm{e}{4}$}
        & \secondcell{$2.305\mathrm{e}{4}$}
        & \bestcell{$2.304\mathrm{e}{4}$} \\
        \hline

        {\sf S3E Tunnel1}
        & $607$ & $835$ & $9.300\mathrm{e}{4}$
        & \bestcell{$9.301\mathrm{e}{4}$}
        & \bestcell{$9.301\mathrm{e}{4}$}
        & \bestcell{$9.301\mathrm{e}{4}$}
        & \bestcell{$9.301\mathrm{e}{4}$}
        & \bestcell{$9.302\mathrm{e}{4}$}
        & \bestcell{$9.302\mathrm{e}{4}$}
        & \bestcell{$9.302\mathrm{e}{4}$} \\
        \hline
    \end{tabular}%
    }
    \label{tab:se3_chordal_sync_async_all}
\end{table*}
\myParagraph{Synchronous Communication}
Tab.~\ref{tab:se3_chordal_sync_async_all} reports the number of nodes and
edges, reference cost, and the cost achieved by each method after 100 synchronous iterations without communication delay.
In the synchronous setting, DeepCORD achieves the best or tied-best cost on
11 out of 13 datasets. Despite utilizing the same two-hop neighbor information, ROBO results in higher costs compared to DeepCORD. This demonstrates that our learned policy better accelerates convergence and minimize objective residuals. For CORD, we use the parameters reported in the original paper on the standard benchmark datasets. For S3E, we performed additional tuning for CORD and use the best constant setting: \(d=4\), \(m=0.7\), and \(\Delta t=0.7\). 
Although DeepCORD uses the same underlying update rule as CORD, the learned adaptive module automatically adjusts solver parameters across different graph topologies and initialization conditions, resulting in the best overall performance across datasets. The \texttt{Rim} dataset further demonstrates the ability of DeepCORD to generalize across problem scales: despite being trained only on graphs with 500 nodes, DeepCORD achieves the lowest cost among all baselines on a graph with more than
10,000 nodes.

\myParagraph{Asynchronous Communication} 
The right panel of Tab.~\ref{tab:se3_chordal_sync_async_all} reports evaluation results under asynchronous communication with randomized delays and packet drop. 
After 100 asynchronous iterations, DeepCORD achieves the lowest cost on the majority of datasets. 
% In the asynchronous setting, CORD uses a single constant parameter setting, \(d=4\), \(m=0.7\), and \(\Delta t=0.15\), while DJ uses a fixed step size of \(1\) without dataset-specific tuning. 
The improvement is particularly evident on the 
\texttt{Grid}, \texttt{Dormitory1}, \texttt{Library1}, and \texttt{Library2} datasets, where DeepCORD achieves lower cost compared to CORD by leveraging the predicted adaptive parameters.
% CORD attains higher final costs, DeepCORD predicts adaptive parameters that maintain stable convergence under such challenging asynchronous conditions.

\begin{figure}[t]
    \centering
    \includegraphics[width=0.85\textwidth]{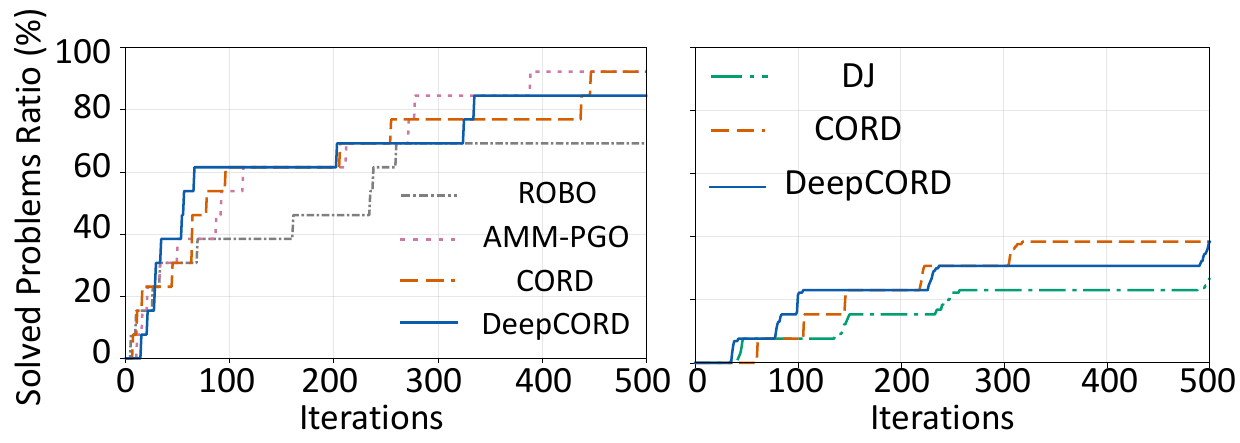}
    \caption{Performance profiles of PGO over 500 iterations ($\Delta = 0.01$) under synchronous and asynchronous communication regimes.}
    \label{fig:profile}
\end{figure}

\myParagraph{Performance Profiles} 
% \YT{For CORD on RSS datasets, use the parameter setting consistent as in RSS paper.}
Following \citep{fan_majorization_2024}, we also report the performance
profile~\citep{dolan2002benchmarking}, which visualizes the percentage of
problems solved by a method as a function of iterations. Given a tolerance
$\Delta$, we define the solved threshold for each problem as $
    \mathcal{C}_{\Delta}
    =
    \mathcal{C}^*
    +
    \Delta
    \left(
        \mathcal{C}^{(0)} - \mathcal{C}^*
    \right)$,
where $\mathcal{C}^{(0)}$ is the cost achieved by initial guess and $\mathcal{C}^*$ is the reference optimal cost. At each iteration, a problem is considered solved by a
method if its cost falls below $\mathcal{C}_{\Delta}$. Fig.~\ref{fig:profile} reports the performance profiles with
$\Delta=0.01$ after 500 iterations. 
The synchronous profile is computed over 13 benchmark sequences, while the asynchronous profile is computed over 130 runs from 10 Monte Carlo trials per sequence. 
Compared to synchronous setting, the asynchronous performance profile reports lower solved ratio due to the additional challenges introduced by communication.
During the first 100 iterations, DeepCORD achieves the highest Area Under the Curve (AUC) compared to all baselines, demonstrating the value of the learned parameter adaptation.
Over the longer horizon, DeepCORD demonstrates stable and competitive convergence across the entire 500 iterations, despite the fact that training only uses $K=50$ unrolled iterations.

% In the synchronous setting, DeepCORD exhibits an Area Under the Curve (AUC) comparable to AMM-PGO, while achieving the highest value in the asynchronous setting.
% 88.48 in the synchronous case, followed by AMM-PGO, CORD, and ROBO, and 70.14 in the asynchronous case, followed by CORD and DJ. 
% The results also show that DeepCORD sustains competitive performance even over longer horizons.

%===============================================================================
\begin{wrapfigure}{r}{0.62\textwidth}
    \centering    \includegraphics[width=0.6\textwidth]{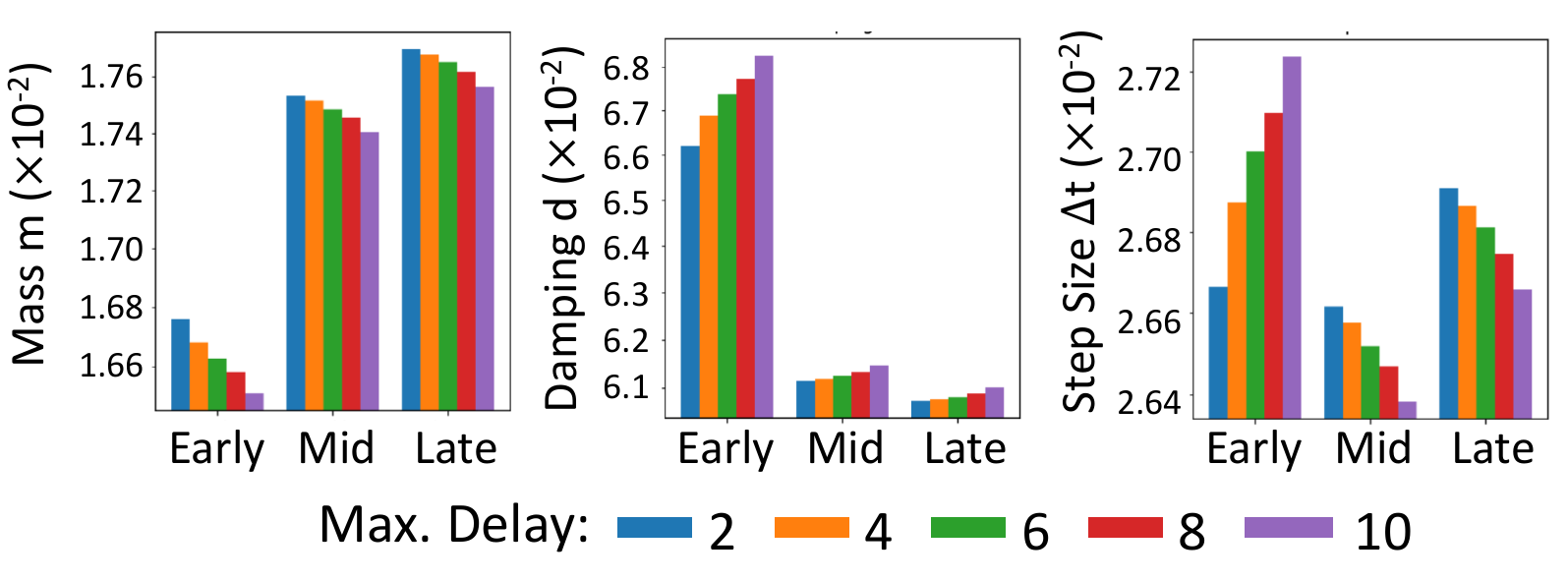}
    \caption{Evolution of the adaptive parameters ($m,d,\Delta t$) predicted by DeepCORD over 100 iterations, illustrated across optimization phases and maximum delays.}
    \label{fig:ParameterAnalysis}
\end{wrapfigure}
\myParagraph{Analysis of Adaptive Parameters}
To provide additional insights, we analyze the DeepCORD parameter predictions on the largest \texttt{Rim} dataset under varied maximum delays ranging from $2$ to $10$ iterations.
The 100 iterations are equally divided into early, middle, and late stages to examine phase-specific behaviors. 

Fig.~\ref{fig:ParameterAnalysis} reports the averaged parameters over 10 Monte Carlo trials per setting. 
Generally, increased delays cause the policy to use more conservative dynamics. 
Specifically, as delay increases, the policy used smaller mass ($m$) and larger damping ($d$) to suppress overshooting and oscillations in the early stage.
In later stages, however, the policy gradually increases mass and decreases damping. 
This indicates that once the solution enters a stable basin, DeepCORD automatically shifts toward accelerated updates. 
The predicted step size $\Delta t$ exhibits more distinct dynamics.
In the early stage, the optimization residuals are amplified by stale information, and the policy uses larger step sizes to reduce residuals. 

As the residual signal weakens, the policy conservatively reduces the step size as the delay increases. 
Finally, in the late stage, the policy increases the step size again to accelerate convergence. 
These results suggest that DeepCORD learns a coupled parameter schedule that flexibly unrolls the effective optimization dynamics according to both the optimization stage and the communication state. 
% \begin{wraptable}{r}{0.46\textwidth}
%     \vspace{-1.0em}
%     \renewcommand{\arraystretch}{1.2}
%     \setlength{\tabcolsep}{0.32em}
%     \centering
%     \caption{SL(4) results on TUM scenarios. \blue{Yulun: the advantage over CORD and DJ is clear. But the margin vs. GTSAM is large. Can we try longer iterations? Otherwise, we need a good discussion about the gap with centralized optimization.}}
%     \resizebox{0.46\textwidth}{!}{%
%     \begin{tabular}{|c||c||c|c|c|}
%         \hline
%         Dataset & Cent. & DeepCORD & DJ & CORD \\
%         \hline
%         {\sf TUM1} & $1.082$ & \cellcolor{green!20}$3.413$ & $6.169$ & \cellcolor{yellow!25}$5.456$ \\
%         \hline
%         {\sf TUM2} & $0.125$ & \cellcolor{green!20}$1.452$ & $2.708$ & \cellcolor{yellow!25}$2.074$ \\
%         \hline
%         {\sf TUM3} & $0.062$ & \cellcolor{green!20}$0.863$ & $2.341$ & \cellcolor{yellow!25}$1.991$ \\
%         \hline
%     \end{tabular}%
%     }
%     \label{tab:sl4_tum_main}
%     \vspace{-1.0em}
% \end{wraptable}

\subsection{Projective Submap Alignment over $\mathrm{SL}(4)$}
%\YT{Start this section by first discussing the qualitative results in Figure 1.}
\begin{wraptable}{r}{0.56\textwidth}
    \renewcommand{\arraystretch}{1.2}
    \setlength{\tabcolsep}{0.30em}
    \centering
    \caption{$\mathrm{SL}(4)$ projective alignment on TUM RGB-D sequences after 100 iterations. Best and second-best methods are highlighted in green and yellow, respectively.}
    \resizebox{0.56\textwidth}{!}{%
    \begin{tabular}{|c||c|c|c||c|c|c|}
        \hline
        Dataset & \# Nodes & \# Edges & Ref. & DeepCORD & DJ & CORD \\
        \hline
        {\sf TUM1}
        & $249$ & $270$
        & $0.876$
        & \cellcolor{green!20}$2.304$
        & $3.822$
        & \cellcolor{yellow!25}$3.342$ \\
        \hline
        {\sf TUM2}
        & $103$ & $111$
        & $0.120$
        & \cellcolor{green!20}$0.777$
        & $1.821$
        & \cellcolor{yellow!25}$1.594$ \\
        \hline
        {\sf TUM3}
        & $89$ & $91$
        & $0.066$
        & \cellcolor{green!20}$1.392$
        & $2.314$
        & \cellcolor{yellow!25}$1.717$ \\
        \hline
    \end{tabular}%
    }
    \label{tab:sl4_tum_main}
\end{wraptable}
In this section, we evaluate DeepCORD over
$\mathrm{SL}(4)$ projective submap alignment as in VGGT-SLAM \cite{maggio2026vggt}. As shown in Fig. \ref{fig:SL4_result}, DeepCORD successfully fuses local dense submaps reconstructed in each session into a globally consistent map under the same asynchronous communication protocol. We simulate a challenging multi-session setting where the factor graph is initialized by coarse alignment from a spanning tree, and inter-session transformations are perturbed with 0.2~rad rotation noise and 0.1~m translation noise. For the baselines, we use the same asynchronous parameter settings as in the PGO experiments. 
Table~\ref{tab:sl4_tum_main} reports the mean cost after 100 iterations over 10 Monte Carlo runs. 
% DeepCORD significantly outperforms CORD and DJ across all three TUM datasets. 
Although a noticeable gap remains relative to the centralized GTSAM baseline due to the severe initial perturbations, DeepCORD significantly outperforms the CORD and DJ baselines.
The results demonstrate that DeepCORD effectively generalizes to Lie groups beyond $\mathrm{SE}(3)$.

\section{Conclusion}
\label{sec:conclusion}
We presented DeepCORD, a learning-augmented framework for distributed factor graph optimization on matrix Lie groups. DeepCORD unfolds CORD into differentiable iterations and learns local feedback policies that adapt solver parameters including momentum, damping, and step size. Through self-supervised training, the resulting solver generalizes across problem instances, optimization stages, and  communication conditions while preserving the structure of the underlying distributed geometric optimizer. Experiments on $\mathrm{SE}(3)$ PGO and $\mathrm{SL}(4)$ projective alignment demonstrate that DeepCORD improves optimization quality and robustness over hand-tuned distributed baselines. These results suggest that learned adaptive dynamics provide a practical path toward scalable and robust distributed geometric optimization.

\section{Limitations}

While DeepCORD outperforms existing distributed baselines, several limitations remain.
First, it does not provide an asymptotic convergence guarantee. A natural direction is to impose explicit stability constraints, for example by restricting predicted parameters to admissible regions of the underlying dynamics as analyzed in \cite{shin2026distributed}. 
Second, due to memory limitations, the current training is restricted to 50 unrolled iterations and thus does not explicitly account for convergence over longer horizons.
Improving the architecture design or training formulation to accelerate long horizon convergence is a valuable future direction.
Third, DeepCORD can be overly conservative on some instances.
For example, on \texttt{CampusRoad3} with synchronous communication, local gradient and residual features become small even though the global error is still large, causing the policy to predict small updates and enter a refinement regime prematurely.
This suggests that the current local features do not fully capture proximity to global optimality.
Future work could incorporate global feedback, such as features from a dual certificate matrix~\cite{tian_distributed_2021}, to distinguish small local gradients from global convergence.

\section*{Acknowledgments}
M. Ghaffari was supported by AFOSR MURI FA9550-23-1-0400 and AFOSR YIP FA9550-25-1-0224.

%===============================================================================

\clearpage
% The acknowledgments are automatically included only in the final and preprint versions of the paper.
% \acknowledgments{If a paper is accepted, the final camera-ready version will (and probably should) include acknowledgments. All acknowledgments go at the end of the paper, including thanks to reviewers who gave useful comments, to colleagues who contributed to the ideas, and to funding agencies and corporate sponsors that provided financial support.}

%===============================================================================

\clearpage
%% Use plainnat to work nicely with natbib. 
{\small
\bibliographystyle{plainnat}
\bibliography{references}
}

\newpage
\appendix
\clearpage
\setcounter{page}{1}
\setcounter{section}{0}

\onecolumn
    \begin{center}
        \Large
        \textbf{Learning Adaptive Solvers for Distributed Factor Graph Optimization on Matrix Lie Groups}    
        \vspace{0.4cm}
        \\
        Appendix
        \vspace{0.8cm} 
    \end{center}
\setcounter{subsection}{0}
\section{Additional Results}

% \red{Move the additional results to the begining of the appendix.}

\subsection{Additional Results for Pose Graph Optimization}

\begin{figure}[!h]
    \centering
\includegraphics[width=1\columnwidth]{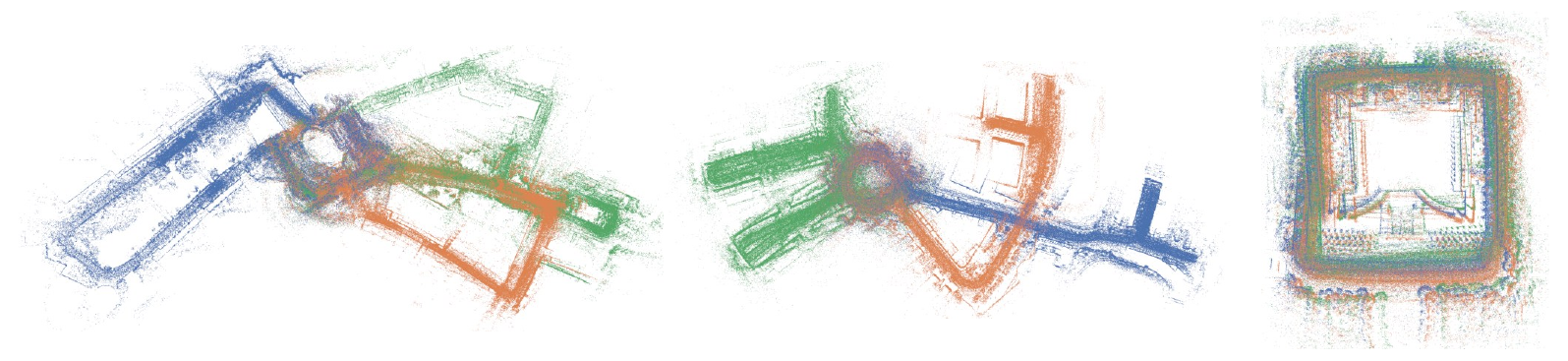}
    \caption{Fused maps from DeepCORD on \texttt{CampusRoad1} (left), \texttt{Dormitory1} (middle), and \texttt{Library1} (right) after PGO. Each color denotes a different robot.} 
    \label{fig:pointcloud}
\end{figure}

This section presents the reconstructed point cloud maps from three representative sequences of the S3E dataset \citeapp{feng2024s3e_app} as additional qualitative results for DeepCORD. As shown in Fig.~\ref{fig:pointcloud}, the submaps obtained from different robots are aligned to recover the global map of each sequence, consistent with the global map provided in \citeapp{feng2024s3e_app}.

\subsection{Additional Results for Projective Submap Alignment}

\begin{figure}[!h]
    \centering
\includegraphics[width=1\columnwidth]{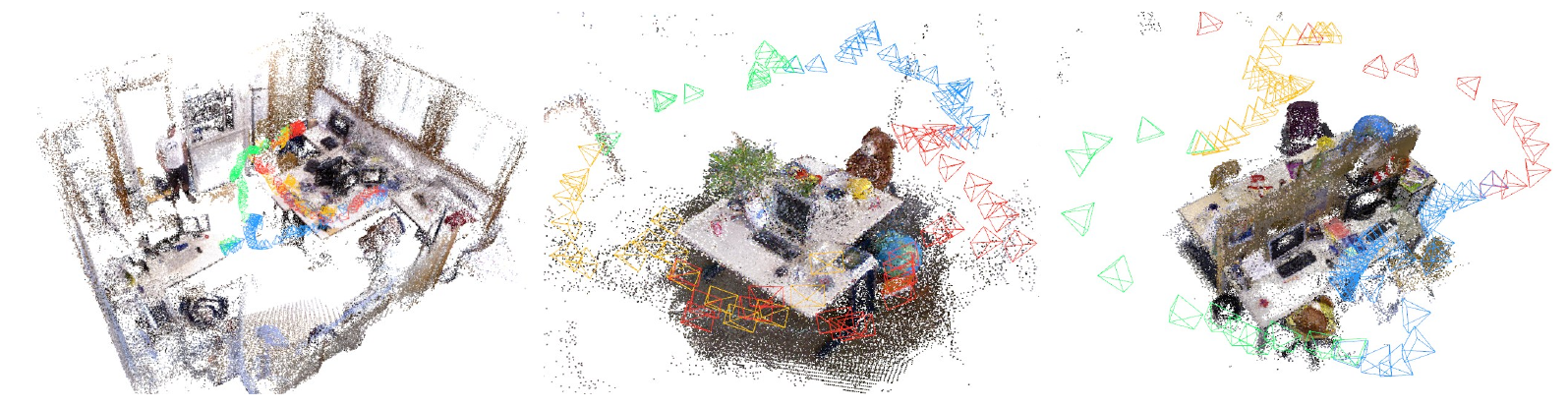}
    \caption{Fused maps from DeepCORD on TUM1 (left), TUM2 (middle), and TUM3 (right) after projective submap alignment over $\mathrm{SL}(4)$. Each colored trajectory denotes a different session.}
    \label{fig:SL4_qual}
\end{figure}

\begin{table*}[!h]
    \renewcommand{\arraystretch}{1.25}
    \setlength{\tabcolsep}{0.25em}
    \centering
    \caption{Evaluation of $\mathrm{SL}(4)$ reconstruction over 10 runs on TUM RGB-D sequences under asynchronous communication. We report accuracy, completion, and Chamfer distance with standard deviations. Best and second-best methods are highlighted in green and yellow, respectively.}
    \resizebox{\textwidth}{!}{%
    \begin{tabular}{|c||c|c|c||c|c|c||c|c|c|}
        \hline
        \multirow{2}{*}{Dataset}
        & \multicolumn{3}{c||}{Accuracy}
        & \multicolumn{3}{c||}{Completion}
        & \multicolumn{3}{c|}{Chamfer} \\
        \cline{2-10}
        & CORD & DJ & DeepCORD
        & CORD & DJ & DeepCORD
        & CORD & DJ & DeepCORD \\
        \hline
        {\sf TUM1}
        & \cellcolor{yellow!25}$0.047\,{\scriptstyle\pm 0.009}$
        & \cellcolor{yellow!25}$0.047\,{\scriptstyle\pm 0.010}$
        & \cellcolor{green!20}$0.045\,{\scriptstyle\pm 0.011}$
        & \cellcolor{yellow!25}$0.043\,{\scriptstyle\pm 0.008}$
        & \cellcolor{yellow!25}$0.043\,{\scriptstyle\pm 0.009}$
        & \cellcolor{green!20}$0.042\,{\scriptstyle\pm 0.009}$
        & \cellcolor{yellow!25}$0.045\,{\scriptstyle\pm 0.008}$
        & \cellcolor{yellow!25}$0.045\,{\scriptstyle\pm 0.009}$
        & \cellcolor{green!20}$0.043\,{\scriptstyle\pm 0.010}$ \\
        \hline
        {\sf TUM2}
        & \cellcolor{yellow!25}$0.052\,{\scriptstyle\pm 0.012}$
        & \cellcolor{green!20}$0.051\,{\scriptstyle\pm 0.011}$
        & $0.053\,{\scriptstyle\pm 0.008}$
        & \cellcolor{yellow!25}$0.050\,{\scriptstyle\pm 0.013}$
        & \cellcolor{green!20}$0.049\,{\scriptstyle\pm 0.012}$
        & $0.052\,{\scriptstyle\pm 0.010}$
        & \cellcolor{yellow!25}$0.051\,{\scriptstyle\pm 0.013}$
        & \cellcolor{green!20}$0.050\,{\scriptstyle\pm 0.011}$
        & $0.052\,{\scriptstyle\pm 0.009}$ \\
        \hline
        {\sf TUM3}
        & \cellcolor{yellow!25}$0.053\,{\scriptstyle\pm 0.006}$
        & \cellcolor{yellow!25}$0.053\,{\scriptstyle\pm 0.007}$
        & \cellcolor{green!20}$0.045\,{\scriptstyle\pm 0.009}$
        & \cellcolor{yellow!25}$0.050\,{\scriptstyle\pm 0.006}$
        & \cellcolor{yellow!25}$0.050\,{\scriptstyle\pm 0.008}$
        & \cellcolor{green!20}$0.045\,{\scriptstyle\pm 0.010}$
        & \cellcolor{yellow!25}$0.052\,{\scriptstyle\pm 0.006}$
        & \cellcolor{yellow!25}$0.052\,{\scriptstyle\pm 0.008}$
        & \cellcolor{green!20}$0.045\,{\scriptstyle\pm 0.009}$ \\
        \hline
    \end{tabular}%
    }
    \label{tab:sl4_tum_reconstruction_metrics_std}
\end{table*}

This section reports additional qualiative results and reconstruction evaluation on
the \texttt{TUM1}, \texttt{TUM2}, and \texttt{TUM3} sequences. Fig.~\ref{fig:SL4_qual} shows the dense
RGB-colored maps obtained after 100 asynchronous iterations on each dataset.
As shown, the maps reconstructed from individual sessions are seamlessly
aligned into a single global map. In addition, we evaluate the reconstructed maps against the maps obtained from
the centralized GTSAM solution using accuracy, completion, and Chamfer distance.
As reported in Tab.~\ref{tab:sl4_tum_reconstruction_metrics_std}, DeepCORD
achieves the best results on the \texttt{TUM1} and \texttt{TUM3} sequences, showing
that the improvement in the optimization objective also results
in improved downstream map quality.

\subsection{Robustness to Initialization Perturbations}
\begin{figure}
    \centering
    \hspace{-0.02\textwidth}
    \includegraphics[width=0.6\textwidth]{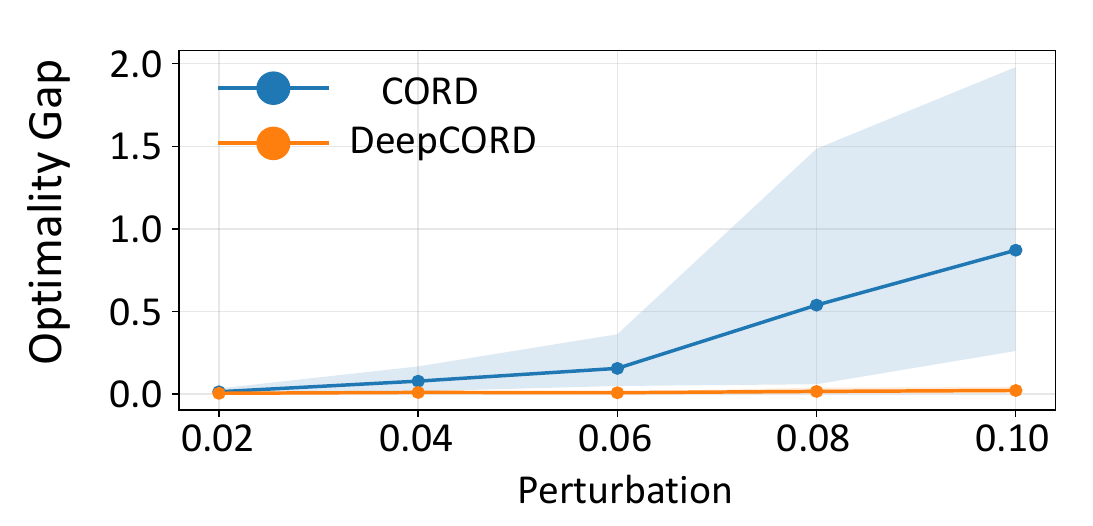}
    \caption{Optimality gap comparison on the \texttt{Rim} dataset under varying initial perturbations. Dots represent the mean over 10 independent runs, and the shaded areas indicate the standard deviation.}
    \label{fig:convergenceBasin}
\end{figure}
This section further evaluates the robustness of DeepCORD under varying initialization quality. Starting from the chordal initialization on the \texttt{Rim} PGO dataset, we perturb the initial estimate with paired random translation and rotation noise \((\sigma\,\mathrm{m}, \sigma\,\mathrm{rad})\), where \(\sigma\in\{0.02, 0.04, 0.06, 0.08, 0.10\}\),
%\YT{0.1 rad is about 5 deg, which is still relatively small perturbation. If there is time left, would be nice to try larger perturbation, e.g., up to 15-20 degree in rotation.} 
and compare CORD and DeepCORD after 100 iterations. For each noise level, the results are averaged over 10 independent Monte Carlo runs. Fig.~\ref{fig:convergenceBasin} reports the resulting optimality gap for each perturbation level. As the perturbation increases, the optimality gap of CORD grows rapidly, indicating that its fixed, hand-tuned parameters are highly sensitive to poor initialization. In contrast, DeepCORD maintains near-optimal convergence across all perturbation levels by adaptively tuning its solver parameters according to the optimization state. 
These results indicate that DeepCORD is substantially less sensitive to initialization perturbations and suggest a wider empirical basin of attraction compared to CORD.
% These results empirically demonstrate that DeepCORD possesses a wider convergence basin than CORD under initialization perturbations.

\subsection{PGO Benchmark Results under Coarse Alignment Initialization}

\begin{table*}[!h]
    \renewcommand{\arraystretch}{1.25}
    \setlength{\tabcolsep}{0.16em}
    \centering
    \caption{SE(3) PGO final cost results from coarse initialization after 100 iterations under synchronous and asynchronous communication. Best and second-best methods are  highlighted in green and yellow, respectively. Diverging cases are denoted as hyphen (---).}
    \resizebox{0.9\textwidth}{!}{%
    \begin{tabular}{|c||c|c|c|c||c|c|c|}
        \hline
        \multirow{2}{*}{Dataset}
        & \multicolumn{4}{c||}{Sync Final Cost}
        & \multicolumn{3}{c|}{Async Final Cost} \\
        \cline{2-8}
        & ROBO & AMM-PGO & CORD & DeepCORD
        & CORD & DJ & DeepCORD \\
        \hline
        {\sf Sphere} & \cellcolor{green!20}$1.687\mathrm{e}{3}$ & \cellcolor{green!20}$1.687\mathrm{e}{3}$ & \cellcolor{yellow!25}$1.688\mathrm{e}{3}$ & \cellcolor{green!20}$1.687\mathrm{e}{3}$ & $2.043\mathrm{e}{3}$ & \cellcolor{yellow!25}$1.765\mathrm{e}{3}$ & \cellcolor{green!20}$1.742\mathrm{e}{3}$ \\
        \hline
        {\sf Torus} & \cellcolor{green!20}$2.423\mathrm{e}{4}$ & \cellcolor{green!20}$2.423\mathrm{e}{4}$ & \cellcolor{yellow!25}$2.426\mathrm{e}{4}$ & \cellcolor{green!20}$2.423\mathrm{e}{4}$ & $2.654\mathrm{e}{4}$ & \cellcolor{yellow!25}$2.484\mathrm{e}{4}$ & \cellcolor{green!20}$2.452\mathrm{e}{4}$ \\
        \hline
        {\sf Grid} & \cellcolor{green!20}$8.432\mathrm{e}{4}$ & \cellcolor{green!20}$8.432\mathrm{e}{4}$ & \cellcolor{yellow!25}$8.456\mathrm{e}{4}$ & \cellcolor{green!20}$8.432\mathrm{e}{4}$ & $9.430\mathrm{e}{4}$ & \cellcolor{yellow!25}$8.559\mathrm{e}{4}$ & \cellcolor{green!20}$8.465\mathrm{e}{4}$ \\
        \hline
        {\sf Cubicle} & $726.137$ & \cellcolor{yellow!25}$720.140$ & $724.054$ & \cellcolor{green!20}$718.757$ & \cellcolor{yellow!25}$1.019\mathrm{e}{3}$ & \cellcolor{green!20}$899.427$ & $1.211\mathrm{e}{3}$ \\
        \hline
        {\sf Rim} & $6.907\mathrm{e}{3}$ & \cellcolor{yellow!25}$5.679\mathrm{e}{3}$ & $5.776\mathrm{e}{3}$ & \cellcolor{green!20}$5.612\mathrm{e}{3}$ & $8.907\mathrm{e}{4}$ & \cellcolor{yellow!25}$9.925\mathrm{e}{3}$ & \cellcolor{green!20}$9.065\mathrm{e}{3}$ \\
        \hline
        {\sf Garage} & \cellcolor{green!20}$1.270$ & \cellcolor{yellow!25}$1.275$ & $1.279$ & $1.301$ & $52.508$ & \cellcolor{green!20}$1.554$ & \cellcolor{yellow!25}$11.942$ \\
        \hline
        {\sf S3E CampusRoad1} & $4.479\mathrm{e}{4}$ & \cellcolor{yellow!25}$4.394\mathrm{e}{4}$ & $4.402\mathrm{e}{4}$ & \cellcolor{green!20}$4.381\mathrm{e}{4}$ & -- & \cellcolor{yellow!25}$5.069\mathrm{e}{4}$ & \cellcolor{green!20}$4.983\mathrm{e}{4}$ \\
        \hline
        {\sf S3E CampusRoad2} & $1.116\mathrm{e}{5}$ & \cellcolor{yellow!25}$1.066\mathrm{e}{5}$ & $1.067\mathrm{e}{5}$ & \cellcolor{green!20}$1.054\mathrm{e}{5}$ & -- & \cellcolor{yellow!25}$1.958\mathrm{e}{5}$ & \cellcolor{green!20}$1.805\mathrm{e}{5}$ \\
        \hline
        {\sf S3E CampusRoad3} & $1.620\mathrm{e}{5}$ & \cellcolor{yellow!25}$1.602\mathrm{e}{5}$ & $1.614\mathrm{e}{5}$ & \cellcolor{green!20}$1.597\mathrm{e}{5}$ & \cellcolor{green!20}$1.793\mathrm{e}{5}$ & $1.839\mathrm{e}{5}$ & \cellcolor{yellow!25}$1.796\mathrm{e}{5}$ \\
        \hline
        {\sf S3E Dormitory1} & $2.052\mathrm{e}{4}$ & \cellcolor{yellow!25}$2.051\mathrm{e}{4}$ & $2.052\mathrm{e}{4}$ & \cellcolor{green!20}$2.047\mathrm{e}{4}$ & -- & \cellcolor{green!20}$2.415\mathrm{e}{4}$ & \cellcolor{yellow!25}$2.495\mathrm{e}{4}$ \\
        \hline
        {\sf S3E Library1} & $3.014\mathrm{e}{4}$ & \cellcolor{yellow!25}$2.995\mathrm{e}{4}$ & $3.024\mathrm{e}{4}$ & \cellcolor{green!20}$2.987\mathrm{e}{4}$ & \cellcolor{yellow!25}$3.154\mathrm{e}{4}$ & $3.213\mathrm{e}{4}$ & \cellcolor{green!20}$3.109\mathrm{e}{4}$ \\
        \hline
        {\sf S3E Library2} & \cellcolor{yellow!25}$2.299\mathrm{e}{4}$ & \cellcolor{green!20}$2.298\mathrm{e}{4}$ & $2.304\mathrm{e}{4}$ & \cellcolor{green!20}$2.298\mathrm{e}{4}$ & \cellcolor{yellow!25}$2.379\mathrm{e}{4}$ & $2.394\mathrm{e}{4}$ & \cellcolor{green!20}$2.324\mathrm{e}{4}$ \\
        \hline
        {\sf S3E Tunnel1} & $9.614\mathrm{e}{4}$ & \cellcolor{yellow!25}$9.518\mathrm{e}{4}$ & $9.644\mathrm{e}{4}$ & \cellcolor{green!20}$9.514\mathrm{e}{4}$ & \cellcolor{green!20}$1.111\mathrm{e}{5}$ & \cellcolor{yellow!25}$1.158\mathrm{e}{5}$ & $1.166\mathrm{e}{5}$ \\
        \hline
    \end{tabular}%
    }
\label{tab:se3_coarse_cost_sync_async}
\end{table*}

To evaluate the robustness of DeepCORD under another initialization scheme, we also report the final cost after 100 iterations using coarse alignment initialization as in ~\citeapp{tian2022kimera_app}. In this setting, we first optimize each robot's individual pose graph and then initialize the multi-robot pose graph using a robot-level spanning tree. Tab.~\ref{tab:se3_coarse_cost_sync_async} shows that DeepCORD achieves the best performance on $20$ out of $26$ sequences, comparable to the result of chordal initialization. Notably, CORD uses the same parameters as in the chordal initialization experiments but diverges on \texttt{CampusRoad1}, \texttt{CampusRoad2}, and \texttt{Dormitory1} in the asynchronous setting, which indicates that a fixed parameter setting fails to adapt to dataset-dependent optimization landscapes. In contrast, DeepCORD consistently reduces the cost by dynamically adjusting its parameters according to the optimization and communication status.

\subsection{Runtime Analysis}
\begin{table}[!h]
    \renewcommand{\arraystretch}{1.25}
    \setlength{\tabcolsep}{0.55em}
    \centering
    \caption{Per-iteration runtime under asynchronous communication. We report the mean runtime per iteration in milliseconds (ms) after 100 iterations. The fastest and second-fastest methods are highlighted in green and yellow, respectively.}
    \resizebox{0.5\columnwidth}{!}{%
    \begin{tabular}{|c||c|c|c|}
        \hline
        Dataset & CORD & DJ & DeepCORD \\
        \hline
        {\sf Sphere}
        & \cellcolor{yellow!25}$37$
        & \cellcolor{green!20}$5$
        & $40$ \\
        \hline
        {\sf Torus}
        & $68$
        & \cellcolor{green!20}$11$
        & \cellcolor{yellow!25}$65$ \\
        \hline
        {\sf Grid}
        & $207$
        & \cellcolor{green!20}$36$
        & \cellcolor{yellow!25}$205$ \\
        \hline
        {\sf Cubicle}
        & $197$
        & \cellcolor{green!20}$31$
        & \cellcolor{yellow!25}$172$ \\
        \hline
        {\sf Rim}
        & \cellcolor{yellow!25}$228$
        & \cellcolor{green!20}$43$
        & $233$ \\
        \hline
        {\sf Garage}
        & $307$
        & \cellcolor{green!20}$52$
        & \cellcolor{yellow!25}$99$ \\
        \hline
        {\sf S3E CampusRoad1}
        & $74$
        & \cellcolor{green!20}$10$
        & \cellcolor{yellow!25}$42$ \\
        \hline
        {\sf S3E CampusRoad2}
        & $159$
        & \cellcolor{green!20}$24$
        & \cellcolor{yellow!25}$76$ \\
        \hline
        {\sf S3E CampusRoad3}
        & $171$
        & \cellcolor{green!20}$26$
        & \cellcolor{yellow!25}$79$ \\
        \hline
        {\sf S3E Dormitory1}
        & $43$
        & \cellcolor{green!20}$6$
        & \cellcolor{yellow!25}$30$ \\
        \hline
        {\sf S3E Library1}
        & $58$
        & \cellcolor{green!20}$8$
        & \cellcolor{yellow!25}$37$ \\
        \hline
        {\sf S3E Library2}
        & $24$
        & \cellcolor{green!20}$4$
        & \cellcolor{yellow!25}$21$ \\
        \hline
        {\sf S3E Tunnel1}
        & $39$
        & \cellcolor{green!20}$6$
        & \cellcolor{yellow!25}$27$ \\
        \hline
  
    \end{tabular}%
    }
    \label{tab:async_runtime_per_iteration}
\end{table}

To quantify the practical overhead, we report the average per-iteration runtime for the asynchronous PGO experiments. We restrict our comparison to CORD and DJ, as the other baselines are implemented in C++. As shown in Tab.~\ref{tab:async_runtime_per_iteration}, DJ yields the lowest runtime since it requires only the computation of the Hessian inverse, whereas CORD and DeepCORD necessitate additional matrix operations associated with integration.

For DeepCORD, the majority of the runtime is consumed by the neural network inference and solving the linear equations involving the Hessian. However, compared to CORD, the efficiency gain achieved by employing a sparse PCG solver more than offsets the additional neural network overhead, leading to faster runtimes across most datasets. Specifically, on the S3E datasets where the pose graph is partitioned among only three robots, CORD's runtime increases due to the dense Cholesky decomposition of large local systems, whereas DeepCORD avoids this direct factorization by employing the sparse PCG solver.

\section{Network Architecture Details}

This section provides additional implementation details of the DeepCORD policy network. 
Recall from Fig.~\ref{fig:architecture} that at iteration $k$, robot $i$ constructs an augmented graph $\mathcal{G}^i_k=(\mathcal{V}^i_k,\mathcal{E}^i_k)$ using its local variables and the states received from neighboring robots. The policy takes node, edge, and graph-level attributes from $\mathcal{G}^i_k$ as input and predicts the adaptive CORD parameters for robot $i$.

\subsection{Model Inputs}
% \YT{We should call these ``attributes'' instead of ``features'' to stay consistent.}

\noindent\textbf{Edge Attributes.}
For each edge $(u,v)\in\mathcal{E}^i_k$, we use the current chordal residual, the measurement precision, and the previous graph-level context as edge features. The chordal residual is computed from
\[
    E^{uv}_k = X^u_k Z^{uv} - X^v_k \in \mathbb{R}^{4\times 4},
\]
where $X^u_k$ and $X^v_k$ are the current pose estimates and $Z^{uv}$ is the relative pose measurement. We vectorize the rotational block of $E^{uv}_k$ and concatenate it with the translational block, resulting in a 12-dimensional residual vector. This residual is scaled by two scalar precision weights, corresponding to the rotational and translational parts. Finally, we concatenate this vector with the two scalar weights and the previous graph-level context vector of dimension $h$, yielding the $(14+h)$-dimensional vector.

\medskip
\noindent\textbf{Node Attributes.}
For each node $u \in \mathcal{V}^i_k$, we construct a node attribute vector comprising the current body velocity, the body-trivialized local gradient, and a scalar communication feature. 
Note that both the body velocity and the gradient are $6$-dimensional vectors in the Lie algebra $\mathfrak{se}(3)$.
The communication feature $c^u_k$ is defined according to the communication status of node $u$:
\[
    c^u_k =
    \begin{cases}
        0, & \text{if } u \text{ is a local node in robot $i$'s own trajectory}, \\
        1, & \text{if } u \text{ is received from a neighbor robot $j$ under synchronous comm.}, \\
        q \, \Delta t^j_{k-q}, & \text{if } u \text{ is received from neighbor } j \text{ under asynchronous comm.},
    \end{cases}
\]
where $q$ denotes the communication delay in iterations and $\Delta t^j_{k-q}$ is the step size included in the delayed packet from robot $j$. 
In the asynchronous case, this feature provides a scalar estimate of the temporal staleness of the received node state. 
Consequently, the total node attribute dimension is $6 + 6 + 1 = 13$.

\medskip
\noindent\textbf{Graph-level Attributes.}
In addition to node and edge attributes, we compute four scalar graph-level statistics that summarize the current optimization state of robot $i$. These are the root-mean-square (RMS) weighted residual over intra-robot edges, the RMS weighted residual over inter-robot edges, the RMS local gradient magnitude over local nodes, and the RMS update magnitude over local nodes, yielding the $4$-dimensional vector.
\subsection{Feature Embedding}

All raw node, edge, and graph-level attributes are transformed using $\operatorname{symlog}$ to reduce the effect of large input magnitudes. The transformed features are then embedded into a latent space of dimension $h=128$. The $13$-dimensional node attribute is linearly projected into the latent space using an MLP. We also add an $8$-dimensional random-walk structural encoding \citeapp{dwivedi2021graph_app} to each node embedding, which provides local topological information of the augmented graph. The $(14+h)$-dimensional edge attribute is also mapped to the latent space using a two-layer MLP and the graph-level attributes are separately processed by another two-layer MLP.

\subsection{GPS Backbone}

The embedded graph is processed by a stack of four GPS layers \citeapp{rampavsek2022recipe_app}. Each GPS layer combines local message passing with global multi-head attention, allowing the policy to use both local graph connectivity and long-range information over the augmented graph. We use 8 attention heads and hidden dimension 128 throughout the GPS backbone. 
The reader is referred to \citeapp{rampavsek2022recipe_app}  for detailed descriptions of the GPS architecture.

\subsection{Parameter Prediction Head}
After the final GPS layer, the node and edge embeddings are mean-pooled over $\mathcal{V}^i_k$ and $\mathcal{E}^i_k$, respectively. These pooled embeddings are then concatenated with the processed graph-level features to form the input for the parameter head of robot $i$ at iteration $k$. This parameter-head MLP yields three unconstrained scalar outputs, \(o^i_{m,k}\), \(o^i_{d,k}\), and \(o^i_{\Delta t,k}\), which correspond to the raw predictions for the mass, damping, and step-size parameters. We map these raw outputs to valid ranges by
\[
    m^i_k = \operatorname{softplus}(o^i_{m,k}) + 10^{-2}, \quad
    d^i_k = \operatorname{softplus}(o^i_{d,k}) + 10^{-2}, \quad
    \Delta t^i_k = \sigma(o^i_{\Delta t,k}).
\]
% \YT{Define the inputs to the softplus and sigmoid above.}

\section{Training Details}
\label{app:training}

\subsection{Training Data Curation}
For the synthetic graphs, each instance is partitioned into $4$ distinct subgraphs, resulting in exactly $125$ poses per robot. Full trajectories from the NCLT dataset are segmented into $3$ to $5$  subgraphs. Real-trajectory graphs are rescaled and downsampled to keep the total graph size close to $500$ poses. If the total number of keyframes \(N_{\mathrm{raw}}\) is already within the range \([450,550]\), no rescaling is applied. Otherwise, we choose a downsampling stride \(s=\lceil N_{\mathrm{raw}}/500\rceil\), uniformly scale the trajectory translations by \(1/s\), and keep every \(s\)-th keyframe within each continuous segment while preserving the final keyframe of the segment. The scale factor is lower-bounded by \(0.05\) for very long trajectories. This keeps the graph size close to \(500\) poses while preventing downsampling from artificially increasing the relative motion scale between consecutive keyframes.

To prevent optimization graphs from being dominated by overly dense duplicate loop closures, we apply a randomized sampling strategy. For real-trajectory graphs, loop closure edges are sampled between pose pairs whose ground-truth distance is below $5\,\mathrm{m}$, with an acceptance probability of $40\%$. For synthetic \texttt{grid} and \texttt{helix} graphs, nearby candidates are defined by a $1.4\,\mathrm{m}$ distance threshold; intra-robot loop closures are accepted with a probability of $20\%$, while inter-robot bridge edges are accepted with a $30\%$ probability. For synthetic \texttt{torus}, \texttt{sphere}, and \texttt{random walk} graphs, the nearby threshold is sampled uniformly from $6.5$ to $7.5\,\mathrm{m}$, capping the density at a maximum of $15$ intra-robot loop closures per robot and $3$ inter-robot bridge edges per robot pair. 

All relative edge measurements are corrupted with zero-mean Gaussian noise whose
standard deviation is proportional to the relative motion magnitude. For an edge
with relative translation length \(d\), the translation noise standard deviation
is set to \(0.05d\,\mathrm{m}\), and the rotation noise standard deviation is
set to \(0.02d\,\mathrm{rad}\). The precision weights are then computed from the
corresponding inverse variances. The initial node states are further perturbed
with Gaussian noise using a standard deviation of \(0.5\,\mathrm{m}\) for
translation and \(0.2\,\mathrm{rad}\) for rotation.

\subsection{Training Hyperparameters}
The model is trained using the Adam optimizer with a constant learning rate of $5\times10^{-5}$ and a batch size of $16$. Training is early-stopped if the validation loss fails to decrease for $3$ consecutive epochs. Tab. \ref{tab:Hyperparameter} further reports additional hyperparameters for training loss, which include regularization weights and damping decaying schedule.

% \red{We are missing some hyperparameters here. For example:
% \begin{itemize}
%     \item $\lambda_b, \lambda_d$ in eq (7)
%     \item $\beta, \tau$ used in the damping regularization term
%     \item Double check main paper for any other hyperparameters
% \end{itemize}
% Similar to Appendix A, report these hyperparameters in a table.
% }
\begin{table}[!h]
    \centering
    \renewcommand{\arraystretch}{1.15}
    \setlength{\tabcolsep}{0.55em}
    \caption{Training hyperparameters for training loss.}
    \label{tab:training_hyperparams_sync_feedback}
    \begin{tabular}{lcc}
        \toprule
        Hyperparameter & Symbol & Value
        \\
        \midrule
        Monotonicity regularization weight & \(\lambda_b\) & \(1.0\) \\
        Damping regularization weight & \(\lambda_d\) & \(5.0\) \\
        Damping decay rate & \(\beta\) & \(0.5\) \\
        Damping decay offset & \(\tau\) & \(5.0\) \\
        \bottomrule
    \end{tabular}\label{tab:Hyperparameter}
\end{table}

\subsection{Implicit Differentiation of PCG Solver}
\label{appendix:implicit_diff}

Unfolding CORD iterations requires backpropagation through the linear
solver involving the local Hessian in \eqref{eq:cord_discrete_acc}. Naively
backpropagating through all PCG inner iterations incurs a prohibitive memory
footprint. Instead, we compute analytical gradients using implicit differentiation. The following result is analogous to prior work, e.g., \citeapp{saravanos2025deep_app}, and is presented here for completeness.

% \textbf{Lemma 1 (Implicit Function Theorem) \citeapp{krantz2002implicit}.} 
% Let \(f : \mathbb{R}^m \times \mathbb{R}^n \rightarrow \mathbb{R}^n\)
% be a continuously differentiable function. Let \((x_0,z_0)\) be a point
% such that \(f(x_0,z_0)=0\). If the Jacobian matrix
% \(\frac{\partial f}{\partial z}(x_0,z_0)\) is invertible, then there
% exists a continuously differentiable function \(z^\ast(\cdot)\), defined
% in a neighborhood of \(x_0\), such that \(z^\ast(x_0)=z_0\) and
% \(f(x,z^\ast(x))=0\). Its derivative is given by
% \begin{equation}
%     \frac{\partial z^\ast}{\partial x}(x)
%     =
%     -
%     \left(
%         \frac{\partial f}{\partial z}(x,z^\ast(x))
%     \right)^{-1}
%     \frac{\partial f}{\partial x}(x,z^\ast(x)).
% \end{equation}

% % Using Lemma 1, we establish the analytical gradients for our PCG solver
% % as follows. Denoting 
% % \YT{Before presenting the theorem, discuss more why the setup in (9) is sufficient for us. Here, we can refer back to (4) and explain how (4) and (9) are related.}

% \YT{Theorem 1 proof does not use Lemma 1. We should either make the connection explicit, or remove Lemma 1.}

At every iteration, DeepCORD unrolls the following simplified ODE  
\begin{equation}
     M\dot{\xi} =  - \operatorname{grad}_{X} \mathcal{C}(X) 
    -D\xi.
    \label{eq:simpleODE}
\end{equation} Denoting right-hand side of \eqref{eq:simpleODE} as a vector $b$, 
we consider solving a general linear system $Mz = b$ using PCG, where $M$ is symmetric positive-definite.
In DeepCORD, $M$ is computed from the Levenberg Marquardt approximation of the local Hessian, and hence is symmetric positive-definite by construction.
The following result establishes the analytical gradients required for deep unfolding.

% \red{
% Update the theorem and proof below so that the gradient wrt M is symmetric.
% i.e., $\nabla_M \mathcal{L}=- \frac{1}{2}v(z^\ast)^\top - \frac{1}{2} z^\ast v^\top.$
% }
% the Gauss--Newton approximiation of block-diagonal Hessian used in the CORD update. We assume that \(M\) is
% symmetric positive definite, which is obtained in practice by adding damping
% and anchor regularization to the positive semi-definite Gauss--Newton
% Hessian. 

\textbf{Theorem 1 (Implicit Differentiation of $Mz=b$).}
Let \(z^\ast \in \mathbb{R}^N\) be the unique solution to the linear system \(M z^\ast = b\), where \(M\in\mathbb{R}^{N\times N}\) is symmetric positive-definite. Given the backward gradient \(g_z := \nabla_{z^\ast}\mathcal{L}\) with respect to a loss function $\mathcal{L}$, let \(v\in\mathbb{R}^N\) be the solution to \(M v = g_z\). Then, the analytical gradients of \(\mathcal{L}\) with respect to \(b\) and \(M\) are:
\begin{equation}
    \nabla_b \mathcal{L} = v,
    \qquad
    \nabla_M \mathcal{L} = -\frac{1}{2} \left( v(z^\ast)^\top + z^\ast v^\top \right).
\end{equation}

\textit{Proof.}
Differentiating the forward relation \(M z^\ast - b = 0\) yields \(M\,dz^\ast + dM\,z^\ast - db = 0\). Solving for \(dz^\ast\) gives:
\begin{equation}
    dz^\ast = M^{-1}(db - dM\,z^\ast) .
\end{equation}
The total differential of the loss \(\mathcal{L}\) is then expanded as:
\begin{equation}
    d\mathcal{L} = g_z^\top dz^\ast = g_z^\top M^{-1} db - g_z^\top M^{-1} dM\,z^\ast .
\end{equation}
Defining \(v\) via \(Mv = g_z\), we substitute \(v^\top = g_z^\top M^{-1}\) to obtain:
\begin{equation}
    d\mathcal{L} = v^\top db - v^\top dM\,z^\ast .
\end{equation}
Using the trace identity \(v^\top dM\,z^\ast = \operatorname{tr}(z^\ast v^\top dM)\) and accounting for the symmetric structure of \(M\) (i.e., \(dM = dM^\top\)), this simplifies to:
\begin{equation}
    d\mathcal{L} = v^\top db - \operatorname{tr}\left( \frac{1}{2}\left(z^\ast v^\top + v(z^\ast)^\top\right) dM \right) .
\end{equation}
Identifying the coefficients of \(db\) and \(dM\) directly yields the results. \(\blacksquare\)

\section{Energy Dissipation of the Simplified Dynamics}
This section justifies our use of the simplified CORD dynamics during training.
Specifically, we will show that the simplified dynamics also
preserves the energy dissipation property in the original CORD paper \citeapp{shin2026distributed_app}. 
Recall that the full CORD dynamics includes the co-adjoint and time-varying
mass terms,
\begin{equation*}
    M\dot{\xi} = -\operatorname{grad}_{X}\mathcal C(X) - D\xi + \operatorname{ad}^{*}_{\xi}(M\xi) - \dot M\xi .
\end{equation*}
Instead, we use a simplified ODE that omits the co-adjoint and
time-dependent mass terms:
\begin{equation}
    M\dot{\xi} = -\operatorname{grad}_{X}\mathcal C(X) - D\xi
    \label{eq:simplified_continuous_dynamics}
\end{equation} where $M\succ 0$ and $D\succeq 0$ are assumed to be constant.

We define the total energy as
\begin{equation*}
     E(X,\xi) = \mathcal C(X) + \frac12 \langle \xi, M\xi\rangle .
\end{equation*}
The first term is the optimization objective as potential energy, and
the second term is the kinetic energy induced by the mass matrix $M$. Since
the gradient is body-trivialized, the time derivative of the objective along
the trajectory satisfies
\begin{equation*}
    \frac{d}{dt}\mathcal C(X(t)) = \left\langle \operatorname{grad}_{X}\mathcal C(X), \xi \right\rangle .
\end{equation*}
Furthermore, by chain rule, the kinetic energy derivative is
\begin{equation*}
    \frac{d}{dt} \left( \frac12 \langle \xi,M\xi\rangle \right) = \langle M\dot{\xi},\xi\rangle .
\end{equation*}
Substituting Eq.~\eqref{eq:simplified_continuous_dynamics} gives %\YT{The following eq uses a different notation for the energy function $E$.}
\begin{equation*}
\begin{aligned}
    \dot{E} &= \left\langle \operatorname{grad}_{X}\mathcal C(X), \xi \right\rangle + \langle M\dot{\xi},\xi\rangle \\
    &= \left\langle \operatorname{grad}_{X}\mathcal C(X), \xi \right\rangle + \left\langle -\operatorname{grad}_{X}\mathcal C(X)-D\xi, \xi \right\rangle \\
    &= -\langle \xi,D\xi\rangle \le 0 .
\end{aligned}
\end{equation*}

% Consequently, under the  positive definite damping, the simplified dynamics dissipates the total
% energy and any limiting point of the simplified
% continuous-time dynamics satisfies
% \begin{equation*}
%     \xi = 0, \qquad \operatorname{grad}_{X}\mathcal C(X)=0 .
% \end{equation*}
Thus, the continuous-time simplified dynamics retains the same qualitative
energy dissipation behavior as CORD and converges to the set of first-order
critical points.

\makeatletter
\setcounter{NAT@ctr}{0}
\@ifundefined{c@NAT@ctr@app}{}{\setcounter{NAT@ctr@app}{0}}
\makeatother

{\small
\bibliographystyleapp{unsrtnat}
\bibliographyapp{appendix_references}
}
\end{document}